\documentclass[final]{neus2025}

\title[Neuro-Symbolic Generative Diffusion Models]
    {Neuro-Symbolic Generative Diffusion Models for Physically Grounded, Robust, and Safe Generation}

\usepackage{times}
\usepackage{enumerate}
\usepackage{enumitem}
\usepackage{bm}
\usepackage{wrapfig}
\usepackage{setspace}
\usepackage[font=small,skip=2pt]{caption}

\usepackage{amsmath}
\usepackage{amssymb}
\usepackage{mathtools}
\usepackage{algorithm}
\usepackage{algpseudocode}
\usepackage{xcolor}

\usepackage{multirow}
\usepackage{multicol}
\usepackage{booktabs}

\usepackage{xcolor} 
\definecolor{myblue}{HTML}{285FB5}

\newcommand{\ra}[1]{\renewcommand{\arraystretch}{#1}}

\newcommand{\bcol}[1]{\textbf{#1}}

\newcommand{\rcol}[1]{\textcolor{purple}{#1}}
\newcommand{\bluecol}[1]{\textcolor{myblue}{#1}}

\DeclareMathOperator*{\argmax}{argmax}
\DeclareMathOperator*{\argmin}{argmin}
\DeclareMathOperator*{\minimize}{Minimize}

\newcommand*{\defeq}{\stackrel{\text{def}}{=}}

\coltauthor{\Name{Jacob~K. Christopher} \Email{csk4sr@virginia.edu}\\
 \Name{Michael Cardei} \Email{ntr2rm@virginia.edu}\\
 \Name{Jinhao Liang} \Email{njs4nu@virginia.edu}\\
 \Name{Ferdinando Fioretto} \Email{fioretto@virginia.edu}\\
 \addr Department of Computer Science, University of Virginia}

\begin{document}

\maketitle

\begin{abstract}
  Despite the remarkable generative capabilities of diffusion models, their integration into safety-critical or scientifically rigorous applications remains hindered by the need to ensure compliance with stringent physical, structural, and operational constraints. To address this challenge, this paper introduces \emph{Neuro-Symbolic Diffusion} (NSD), a novel framework that interleaves diffusion steps with symbolic optimization, enabling the generation of certifiably consistent samples under user-defined functional and logic constraints.
  This key feature is provided for both standard and discrete diffusion models, enabling, for the first time, the generation of both continuous (e.g., images and trajectories) and discrete (e.g., molecular structures and natural language) outputs that comply with constraints. 
  This ability is demonstrated on tasks spanning three key challenges: \emph{(1)~Safety}, in the context of non-toxic molecular generation and collision-free trajectory optimization; \emph{(2)~Data scarcity}, in domains such as drug discovery and materials engineering; and \emph{(3)~Out-of-domain generalization}, where enforcing symbolic constraints allows adaptation beyond the training distribution. 
\end{abstract}


\section{Introduction}

\emph{Diffusion models} \citep{ho2020denoising} are a class of generative AI models at the forefront of high-dimensional data creation and form the backbone of many state-of-the-art image and video generation systems \citep{rombach2022high,betker2023improving,liu2024sorareviewbackgroundtechnology}. 
{This potential has also been recently extended to the context of discrete outputs, which is suitable for language modeling or combinatorial structure design, like chemical compounds or peptide design \citep{zheng2024masked,loudiscrete,shi2024simplified}.}  
Diffusion models operate by progressively introducing controlled random noise into the original content and learning to reverse the process to reconstruct statistically plausible samples.
This approach has shown transformative potential for engineering, automation, and scientific research through applications including generating trajectories for robotic agents in complex, high-dimensional environments or synthesizing new molecular structures with improved strength, thermal resistance, or energy efficiency 
\citep{carvalho2023motion,watson2023novo}.  

However, as opposed to standard image synthesis tasks, scientific applications of diffusion models need to be controlled by precise mechanisms or properties that must be imposed on generations. 
{\em While diffusion models produce statistically plausible outputs, they are simultaneously unable to comply with fundamental physics, safety constraints, or user-imposed specifications \citep{motamed2025generativevideomodelslearn}.} Violations of established principles and constraints not only undermine the utility of generative models, but also erode their trustworthiness in high-stakes domains. 
For example, embodied agents powered by generative AI such as drones or robotic arms are susceptible to adversarial manipulations that bypass safety protocols \citep{robey2024}.
To date, these models have struggled to produce even simple trajectories that satisfy basic collision avoidance, let alone more stringent safety requirements in complex environments \citep{power2023sampling}. Similarly, in scientific and industrial applications such as autonomous bio-labs, systems may improperly model specifications or even react to adversarial triggers, potentially leading to synthesis of hazardous compounds \citep{wittmann2024exploring}. 
\emph{Thus, there is a pressing need for generative models to satisfy physical, operational, and structural constraints that govern large-scale scientific and engineering challenges.}

Recently, \cite{christopher2024constrained} observed that a class of generative models can ground their induced distributions to a specific property. Inspired by this observation, this paper provides a step towards addressing the challenge of constraining generative models and developing a novel integration of symbolic optimization with generative diffusion models. The resulting framework, called \emph{Neuro-Symbolic Diffusion} (NSD), enables the generation of outputs that are \emph{certifiably consistent} with user-defined properties, ranging from continuous constraints, such as structural properties for material science applications or collision avoidance in motion-planning environments, to discrete constraints, including the prevention of toxic substructures in molecule generation tasks. 


\noindent\textbf{Contributions.} The contributions of this study are as follows.
\begin{enumerate}[leftmargin=*, parsep=0pt, itemsep=0pt, topsep=0pt]
\item It develops a novel methodology \emph{for integrating functional and logic constraints within generative diffusion models}. 
The core concept involves a tight integration of differentiable constraint optimization within the reverse steps of diffusion process and ensures that each generated sample respects user-imposed or domain-specific properties. 
\item It shows that this approach is not only viable for generation within continuous subspaces but also effective in constraining token generation for discrete modalities, including domain-specific sequence generation for scientific discovery and open-ended language generation. 
\item It provides theoretical grounding to demonstrate when and why constraint adherence can be certified during the neuro-symbolic generative process. 
\item 
It presents an extensive evaluation across three key challenges: (1) \emph{Safety}, demonstrated through non-toxic molecular generation and collision-free trajectory optimization; (2) \emph{Data scarcity}, with applications in drug discovery and materials engineering; and (3) \emph{Out-of-domain generalization}, where enforcing symbolic constraints enables adaptation beyond the training distribution.
\end{enumerate}

\noindent These advances bring forward two key features that are important for the development of generative models for scientific applications: \emph{Improved assurance}, e.g., the models can implement safety predicates needed in the domain of interest, such as natural language generation where prompts could be engineered to elicit harmful outputs, and \emph{Improved generalization},  e.g., the imposition of knowledge and symbolic constraints dramatically improves the model generalizability across domains. 


\section{Preliminaries: Generative Diffusion Models}
\label{sec:prelim}
Diffusion models define a generative process by learning to reverse a \emph{forward stochastic transformation} that progressively corrupts structured data into noise. The generative model then approximates the inverse of this transformation to restore the original structure, thereby allowing sampling from the learned distribution. 
The forward \emph{noising} process $\{\bm x_t\}_{t=0}^T$ progressively corrupts data in a Markovian process, starting from $\bm{x}_0 \sim p_{\text{data}} (\bm x_0)$ and culminating into noise $\bm{x}_T \sim p(\bm{x}_T)$.  
Here, $p_\text{data}(\bm{x}_0)$ represents the distribution induced by real data samples, and $p(\bm{x}_T)$ is, by design, some known distribution. 
The reverse process starts from $\bm{x}_T \sim p(\bm{x}_T)$ and produces samples $\bm{x}_0$ that follow $p_{\text{data}}$. 

In this study, we consider two settings: (1) \emph{continuous diffusion models} \citep{ho2020denoising,song2020score} for data in $\mathbb{R}^d$ (e.g., images or trajectories), and (2) \emph{discrete diffusion models} \citep{loudiscrete,sahoo2024simple}, which were recently introduced to handle discrete data (e.g., sequences of tokens representing natural language or molecular structures).

\noindent\textbf{Diffusion models for continuous data.}
For data in $\mathbb{R}^d$, the forward diffusion process is often modeled as a \emph{stochastic differential equation (SDE)} of the form  
\(
  d\bm{x}_t = -\frac{1}{2} \beta(t) \bm{x}_t dt + \sqrt{\beta(t)} d\bm{B}(t),
\)
where \(\bm{B}(t)\) denotes standard Brownian motion and $\beta(t)$ defines a noise schedule. As $t \to T$, the process asymptotically transforms data into an isotropic Gaussian distribution. The \emph{reverse process}, which recovers the original data, follows a time-reversed SDE underlying \emph{Langevin dynamics}:
\begin{equation}
\label{eq:sgld_cont}
 d\bm{x}_t = \left[-\frac{1}{2} \beta(t) \bm{x}_t + \nabla_{\bm{x}_t} \log p(\bm{x}_t)\right] dt + \sqrt{\beta(t)} d\bm{B}(t).
\end{equation}
However, since exact integration of this process is intractable, in practice, it is discretized into a \emph{finite-step Markov chain}:  
\begin{equation}
\label{eq:sgld_disc}
\bm{x}_{t - \Delta} = \bm{x}_t + \gamma_t \bm{s}_{\theta}(\bm{x}_t, t) + \sqrt{2\gamma_t} \epsilon,
\end{equation}
where $\bm{s}_{\theta}$ is a neural network that approximates the 
gradient of the log data distribution $\nabla_{\bm{x}_t} \log p(\bm{x}_t)$, called the \emph{score function}, and is used to guide the model toward high-density regions. Additionally, $\gamma_t$ is the step size and $\epsilon$ is a Gaussian perturbation. 
In the deterministic limit (i.e., \(\gamma_t\to 0\)), this becomes a pure gradient ascent update on \(\log p(\bm{x}_t)\).

\noindent \textbf{Discrete diffusion models.}  
For discrete data such as text tokens, each sample is a sequence $\bm{x}_0 = (\bm{x}_0^1, \ldots, \bm{x}_0^{L})$  where each token $\bm{x}_0^i \in \mathbb{R}^V$ is represented as a one-hot vector over a vocabulary of size $V$.
The forward process progressively corrupts the sequence by replacing tokens with noise, 
the marginal of which is defined as:
\(
    q(\bm{x}_t \mid \bm{x}_0) \;=\; 
    \mathrm{Cat}\left(\bm{x}_t;\; (1 - \beta(t)) \,\bm{x}_0 
      \;+\; \beta(t)\, \bm{\nu} \right),
\)
where $\beta(t) \in [0,1]$ is a schedule that increases with $t$, so that tokens are increasingly replaced by noise, and $\mathrm{Cat}(\cdot; \bm{z})$ denotes a categorical distribution parameterized by probability vector $\bm{z} \in \Sigma^{V}$, where $\Sigma^{V}$ denotes the $V$-dimensional simplex.
Finally, $\bm{\nu}$ is a fixed categorical distribution, often concentrated on a special token, such as [MASK] (as in MDLM from \cite{sahoo2024simple}) or chosen uniformly (as in UDLM from \cite{schiff2024simple}). It models a process akin to that induced by the isotropic Gaussian in the continuous counterpart. 
As $t$ increases, each token in $\bm{x}_t$ becomes less correlated with its original value, and approaches the noise distribution. 
The \emph{reverse process} is represented as
\begin{equation}
\label{eq:discrete_reverse}
\bm{x}_{t-\Delta} =
\begin{cases}
\mathrm{Cat}\bigl(\bm{x}_{t-\Delta}; \bm{x}_t\bigr), 
  & \text{if } \bm{x}_t \neq \bm{\nu},\\[4pt]
\mathrm{Cat}\!\Bigl(
  \bm{x}_{t-\Delta}; 
  \frac{\beta(t-\Delta)\,\bm{\nu} + (\beta(t)- \beta(t-\Delta))}{\beta(t)}\, \bm{s}_{\theta}(\bm{x}_t,t)
\Bigr), 
  & \text{if } \bm{x}_t = \bm{\nu},
\end{cases}
\end{equation}
Since \(\bm{x}_0\) is unknown at inference, it is approximated with 
\(\bm{s}_{\theta}(\bm{x}_t, t)\). Here, \(\bm{x}_t\) represents 
a vector of \emph{probability distributions} over tokens at each position 
in the sequence. The paper denotes with $\bm{x}^\star_t = \argmax(\bm{x}_t)$ 
as the selected output sequence, 
where the $\argmax$ operator is applied independently to each member 
$\bm{x}_t^i$ of the sequence $\bm{x}_t$.

\section{Related Work and Limitations}
Despite their success, existing diffusion models struggle to enforce structured constraints.
An approach developed to address this issue relies on sampling a conditional distribution \(p_\text{data}(\bm{x}_0 \mid \mathbf{c})\), where \(\mathbf{c}\) conditions the generation. 
This approach transforms the denoising process via 
\begin{wrapfigure}[12]{r}{0.42\linewidth}
    \vspace{-10pt}
    \centering
    \includegraphics[width=0.99\linewidth]{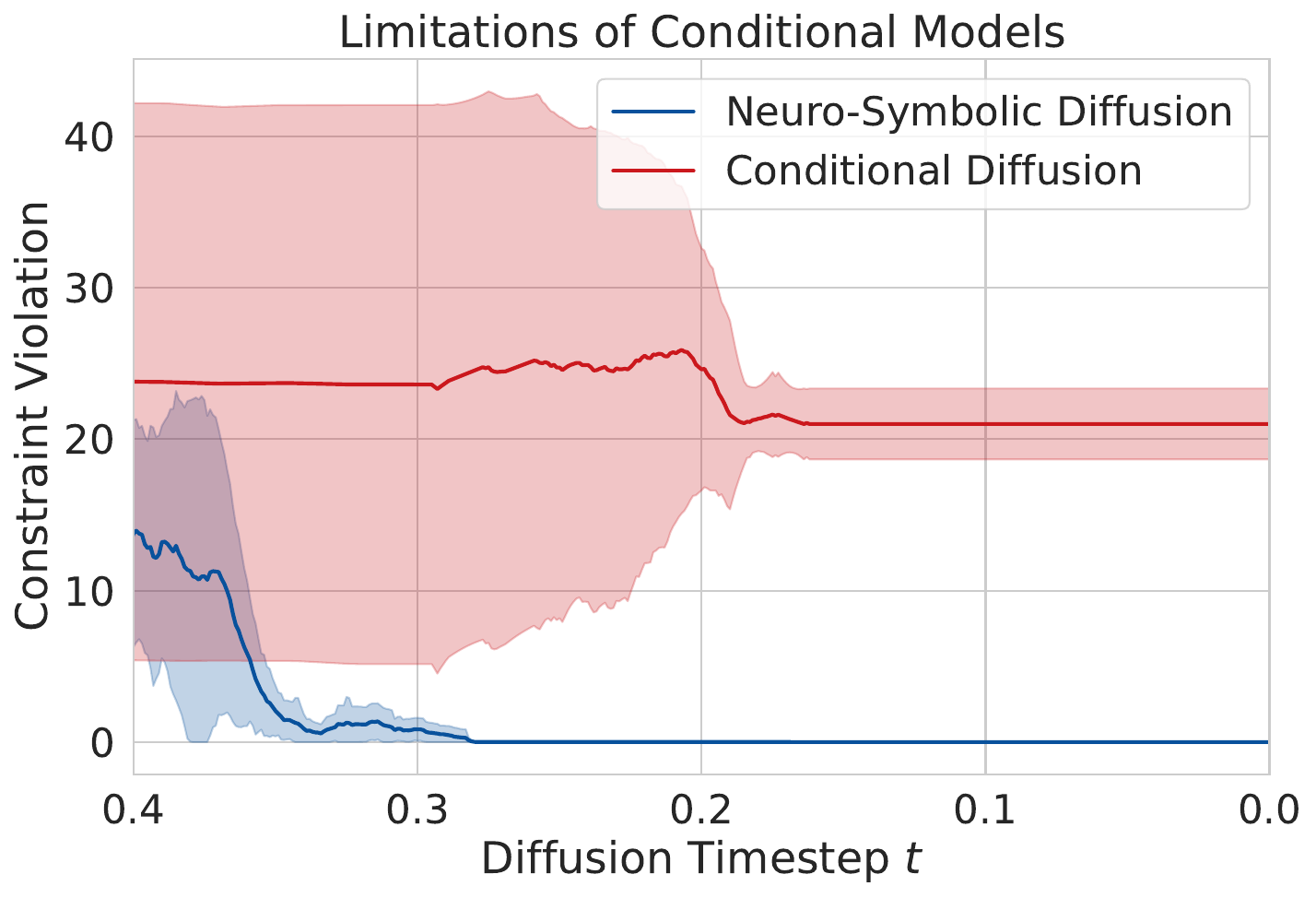}
    \vspace{-12pt}
    \renewcommand{\baselinestretch}{0.9} 
    \caption{\small \rcol{Conditional models} fail to converge to feasible states while \bluecol{Neuro-Symbolic Diffusion} produces no violations.}
    \renewcommand{\baselinestretch}{1.0} 
    \label{fig:constr_conv}
\end{wrapfigure} 
classifier-free guidance:
\[
    \hat{\bm s}_\theta \defeq 
    \lambda \times \bm{s}_\theta(\bm{x}_t, t, \bm{c}) +
    (1-\lambda) \times \bm{s}_\theta(\bm{x}_t, t, \bot),
\]
where $\lambda \in (0,1)$ is the \emph{guidance scale} and $\bot$ is a null vector representing non-conditioning \citep{ho2022classifier}. 
These methods have demonstrated effectiveness in capturing physical design properties \citep{wang2023diffusebot}, positional awareness \citep{carvalho2023motion}, and motion dynamics \citep{yuan2023physdiff}. However, while conditioning can guide the generation process, it offers no reliability guarantees. 
This issue is illustrated in Figure~\ref{fig:constr_conv} (\rcol{red curve}), which shows the magnitude of constraint violations 
observed in a physics-informed motion experiment simulating the dynamics of an object under a force-field influence (more details provided in Appendix \ref{appendix:phys}). Here, the conditional model fails to adhere to motion constraints as the diffusion steps evolve $(t\to 0)$. 
Additionally, conditioning in diffusion models often necessitates the training of auxiliary classification or regression models, which 
requires additional labeled data. This is highly impractical in many scientific contexts of interest to this paper, where sample collection is expensive or extremely challenging. 

An alternative approach involves applying \emph{post-processing steps} to correct deviations from desired constraints in the generated samples. This correction is typically implemented in the last noise removal stage 
\citep{giannone2023aligning, power2023sampling,maze2023diffusion}. However, 
this approach present two main limitations. First, the objective does not align with optimizing the diffusion model score function, and thus does not guide the model towards high-density regions. This inherently positions the diffusion model's role as ancillary, with the final synthesized data often resulting in a significant divergence from the learned (and original) data distributions.
Second, these methods are reliant on a limited and problem specific class of objectives and constraints, such as specific trajectory ``constraints'' or shortest path objectives which can be integrated as a post-processing step \citep{giannone2023aligning, power2023sampling}.

To overcome these gaps and handle arbitrary symbolic constraints, our approach casts the reverse diffusion process as a differentiable constraint optimization problem, which is then solved through the application of repeated projection steps. 
The next section focuses on continuous models for clarity, but this reasoning extends naturally to discrete models, as shown in Appendix~\ref{appendix:concrete}. 


\section{Reverse Diffusion as Constrained Optimization}
\label{sec:rev_opt}
In traditional diffusion model sampling, the reverse process transitions a noisy sample $\bm{x}_T$ to $\bm{x}_0$ by reversing the stochastic differential equation in \eqref{eq:sgld_cont}, which is discretized into an iterative Langevin dynamics update in \eqref{eq:sgld_disc}. 
The key enabler for the integration of constraints in the diffusion process is the realization that \emph{each reverse step can be framed as an optimization problem}. 
As shown in \cite{xu2018global}, under appropriate regularity conditions, Langevin dynamics converges to a stationary distribution $p(\bm{x}_t)$, effectively maximizing $\log p(\bm{x}_t)$ \citep{christopher2024constrained}. As $t \to 0,$ the variance schedule decreases and the noise term $\sqrt{2\gamma_t}\,\bm{\epsilon}$ vanishes, causing the update step to become deterministic gradient ascent on $\log p(\bm{x}_t)$. This perspective reveals that the reverse process can be viewed as \emph{minimizing the negative log-likelihood of the data distribution}. The proposed method for constraining the generative process relies on this interpretation.

In traditional score-based models, at any point throughout the reverse process, $\bm{x}_t$ is unconstrained. When these samples are required to satisfy certain constraints, the objective remains unchanged, but the solution to this optimization must fall within a feasible region $\mathbf{C}$. Thus, the optimization problem formulation becomes:
\begin{equation}
\label{eq:rev_opt}
    \minimize_{\bm{x}_t \;:\; t \in (0,T]}
\quad
\int_{t=0}^{T} -\log p\big(\bm{x}_t \mid \bm{x}_0\big)
\quad
\text{subject to}
\quad
\bm{x}_t \,\in\, \mathbf{C}\,, \;\forall t\in (0,T].
\end{equation}
In practice, \(\mathbf{C}\) is defined by the intersection of multiple ($n$) functional expressions or logic predicates: $\mathbf{C} \defeq \bigwedge_{i=1}^n \phi_i(\bm{x})$, where each $\phi_i(\bm{x})$ is a predicate that returns 1 if \(\bm{x}\) satisfies a condition and 0 otherwise. 
For example, these may be a series of properties that are required for a generated molecule to be a non-toxic chemical compound. The paper uses $\phi_{1:n}$ to denote the subset of $n$ constraints in $\bm{C}$. 

In discrete diffusion models, the score function can be similarly modeled through Concrete Score Matching as expounded on in Appendix~\ref{appendix:concrete} \citep{meng2022concrete}.
Hence, the underlying optimization interpretation remains consistent: each denoising update seeks to move $\bm{x}_t$ closer to the high-density region of the learned distribution while respecting the constraints $\mathbf{C}$.

\section{Neuro-Symbolic Generative Models}

The score network $\bm{s}_\theta(\bm{x}_t, t)$ directly estimates the first-order derivatives of Equation \eqref{eq:rev_opt} (excluding the constraints) and provides the necessary gradients for iterative updates defined in Equations \eqref{eq:sgld_disc} and \eqref{eq:discrete_reverse}. In the presence of constraints, however, an alternative iterative method is necessary to guarantee feasibility. 
This section illustrates how projected guidance can augment diffusion sampling and transform it into a constraint-aware optimization process. 
First, it formalizes the notion of a projection operator $\mathcal{P}_{\mathbf{C}}$, which finds the nearest feasible point to the input $\bm{x}$:
\begin{equation}
    \label{eq:projection}
    \mathcal{P}_{\mathbf{C}}(\bm{x}) \defeq \argmin_{\bm{y}} 
    \begin{cases}
        ||\bm{x} - \bm{y}||_{2}^2 \;\; \;\;\; \; \;\; \; \text{s.t.} \; \bm{y}\in \mathbf{C}, 
        & \text{if } \bm{x} \; \text{is continuous},\\[4pt]
        D_{\mathrm{KL}}\left(\bm{x} \; \| \; \bm y\right) \;\;\;\;  \text{s.t.} \; \bm{y}^\star = \argmax(\bm{y}) \in \mathbf{C}, 
        & \text{if } \bm{x} \; \text{is discrete}.\\[4pt]
    \end{cases}
\end{equation}
Because continuous diffusion operates in a multi-dimensional real space, \(\bm{x} \in \mathbb{R}^d\), whereas discrete diffusion represents samples as \(\bm{x} \in \Sigma^V\), the notion of proximity must be adapted accordingly. In continuous settings, Euclidean distance provides a natural measure of deviation from feasibility.
At the same time, for discrete models, the underlying representations correspond to probability distributions; thus, the Kullback–Leibler (KL) divergence is chosen to quantify the minimal adjustment needed to satisfy the constraints.  
The optimization objective in Equation \eqref{eq:projection} defines a projection operator that minimizes a cost function, which we refer to as the \emph{cost of projection}. In the continuous setting, this corresponds to the squared Euclidean distance, \( \| \bm{y} - \bm{x}\|_{2}^2 \), while in the discrete setting, it is determined by the KL divergence. More generally, we denote this projection cost as \(D_{\mathrm{cost}}(\bm{x}, \bm{y})\), representing the modality-specific distance minimized in the projection step.

To ensure that feasibility is maintained throughout the reverse process of the diffusion model, the sampling step is updated as:
\begin{equation}
    \label{eq:update_step}
    \bm{x}_{t-\Delta} =
    \begin{cases}
        \mathcal{P}_\mathbf{C}\Bigl(\bm{x}_t + \gamma_t \bm{s}_{\theta}(\bm{x}_t, t) + \sqrt{2\gamma_t} \epsilon \Bigr)
        & \text{if } \bm{x} \; \text{is continuous},\\[4pt]
         \mathcal{P}_\mathbf{C}\Bigl(\mathrm{Cat}\bigl(\cdot\,; \pi_\theta(\bm{x}_t,t)\bigr) \Bigr)
        & \text{if } \bm{x} \; \text{is discrete}.\\[4pt]
    \end{cases}
\end{equation}
where \(\gamma_t > 0\) is the step size, \(\epsilon \sim \mathcal{N}(0,I)\), and \(\pi_\theta(\bm{x}_t,t)\) is the predicted probability vector, as generalized from Equation \eqref{eq:discrete_reverse}.
Hence, at each step of the Markov chain, a gradient update is applied to minimize the objective in Equation \eqref{eq:rev_opt}, while interleaved projections ensure feasibility throughout the sampling process. Importantly, convergence is guaranteed for convex constraint sets (see Section \ref{sec:theory}), and empirical results in Section \ref{section:exp} demonstrate the effectiveness of this approach, even in highly non-convex settings. Notably, the projection operators can be warm-started across iterations, providing a practical solution for efficiently handling regions with complex constraints.

\begin{wrapfigure}[12]{r}{0.44\linewidth}
        \vspace{-20pt}
        \begin{minipage}[t]{\linewidth}
         \begin{algorithm}[H]
            \DontPrintSemicolon
            \caption{\footnotesize Augmented Lagrangian Projection}
            \label{alg:alm}
            {\footnotesize
            \KwIn{$\bm{x}_t$, $\lambda$, $\mu$, $\gamma$, $\alpha$, $\delta$}
            
            $\bm{y} \gets \bm{x}_t$
            
            \While{$\tilde{\phi}(\bm{y}) < \delta$}{
                \For{$j \gets 1$ \KwTo max\_inner\_iter}{
                    
                    $\mathcal{L}_{\text{ALM}} \gets D_{\mathrm{cost}}\bigl(\bm{x}_t,\, \bm{y}\bigr) + \lambda\,\tilde{\phi}(\bm{y}) + \tfrac{\mu}{2}\,\tilde{\phi}(\bm{y})^2$
                    
                    $\bm{y} \gets \bm{y} - \gamma\, \nabla_{\bm{y}} \mathcal{L}_{\text{ALM}}$
                    
                }
                $\lambda \gets \lambda + \mu\, \tilde{\phi}(\bm{y}); \;\;
                \mu \gets \min\bigl(\alpha\mu,\, \mu_{\text{max}}\bigr)$\;
                
            }
            $\bm{x}_{t-\Delta} \gets \bm{y}$ 
            
            \Return $\bm{x}_{t-\Delta}$\;
            }
        \end{algorithm}
    \end{minipage}%
    \label{fig:side_by_side_algorithms}
\end{wrapfigure}
\noindent \textbf{Augmented Lagrangian Projection.}  
To solve the projection subproblem \(\mathcal{P}_\mathbf{C}(\bm{x}_t)\) in each sampling step, the paper uses a Lagrangian dual method \citep{boyd2004convex}, 
where the constraints are incorporated into a relaxed objective by using Lagrange multipliers $\lambda$ and a quadratic penalty term \(\mu\). The augmented 
Lagrangian function is defined as
\[
  \mathcal{L}_{\text{ALM}}\bigl(\bm{y}, \lambda, \mu\bigr) \!=\! D_{\mathrm{cost}}\bigl(\bm{x}_t,\,\bm{y}\bigr)\!+\!\lambda\,\tilde{\phi}(\bm{y})\!+\!\tfrac{\mu}{2}\,\tilde{\phi}(\bm{y})^2,
\]
where \(\tilde{\phi}\) denotes a differentiable residual or constraint violation of the original (potentially non-differentiable) constraint function \(\phi_{1:n}\). For example, consider a linear constraint $\phi = A\bm{y} \leq b$, then $\tilde{\phi} = \max(0, A\bm{y}-b)$. 
The iterative update follows a dual ascent strategy, where the variables \(\bm{y}\) are optimized via gradient step on \(\nabla_{\bm{y}} \mathcal{L}_{\text{ALM}}\), while the dual variables \(\lambda\) are updated by 
\(
  \lambda \to \lambda + \mu \tilde{\phi}(\bm y).
\)
Additionally, the penalty coefficients \(\mu\) are increased adaptively by $\alpha$ to tighten constraint enforcement. This procedure continues until \(\tilde{\phi}(\bm{y}) < \delta\) or the maximum iteration count is reached, returning a feasible \(\bm{y}\) as \(\bm{x}_{t-\Delta}\), as illustrated in Algorithm \ref{alg:alm}. 
Note that for convex constraint sets, the augmented Lagrangian method provides strong theoretical guarantees for exact convergence to the projection onto the feasible set \citep{boyd2004convex}. Specifically, if Slater’s condition holds (i.e., there exists a strictly feasible point), then the method guarantees convergence to the primal solution satisfying the constraints. This feature is key for several applications of interest to this work (see Section \ref{section:exp}). 

Notice that, for discrete variables, the projection \(\mathcal{P}_\mathbf{C}(\bm{x}_t)\) must be imposed on the decoded sequence \(\bm{y}^\star = \argmax(\bm{y})\). Because the $argmax$ operator is not differentiable, this paper adopts a Gumbel-Softmax relaxation \citep{jang2017categorical} to preserve gradient-based updates. Details on this relaxation are provided in Appendix~\ref{appendix:gumbel}, and further technical aspects of the augmented Lagrangian scheme are discussed in Appendix~\ref{appendix:alm}.

By incorporating constraints throughout the sampling process, the interim learned distributions are steered to comply with these specifications. 
The effectiveness of this approach is empirically evident from Figure~\ref{fig:constr_conv} (\textcolor{myblue}{blue curve}): remarkably, as the reverse process unfolds, constraint violations steadily approach $0$ 
and a theoretical justification for the validity of this approach is provided in the next section. A key distinction of this method, in contrast to prior approaches \citep{giannone2023aligning, power2023sampling}, is that it {\em optimizes the negative log-likelihood as the primary sampling objective}, maintaining consistency with standard unconstrained diffusion models while enforcing verifiable constraints. This provides a fundamental advantage: \emph{it maximizes the probability of generating samples that conform to the data distribution while ensuring feasibility}. In contrast, existing baselines prioritize external constraints at the expense of distributional fidelity, often leading to significant deviations from the learned distribution, as shown in Section~\ref{section:exp}.


\section{Effectiveness of Neuro-Symbolic Generation: A Theoretical Justification}
\label{sec:theory}

This section focuses on two key outcomes of incorporating iterative projections during diffusion sampling:  
\textbf{(1)} As the sample \(\bm{x}_t\) transitions toward the minimizer of the negative log-likelihood (the primary diffusion objective), each projection step needs only a small adjustment to maintain feasibility. Thus, the projected sampling remains closely aligned with the unconstrained score-based dynamics, causing minimal deviation from the main objective.  
\textbf{(2)} By keeping the sample near \(\mathcal{P}_\mathbf{C}(\bm{x}_t)\) throughout the sampling trajectory, any subsequent or “final” projection step becomes less costly (e.g., smaller Euclidean or KL distance).  

Proofs for all theorems are provided in Appendix \ref{appendix:theory} and additional technical details in Appendix \ref{appendix:concrete}.
The analysis assumes a convex feasible region \(\mathbf{C}\) and unifies results for both continuous (Euclidean) and discrete (KL-based) metrics \citep{christopher2024constrained}. Below, we detail the theoretical underpinnings using the update notation \(\bm{x}_t \to \bm{x}_{t-\Delta}\) in place of traditional iterative indexing.

\noindent Consider an update step that transforms a sample \( \bm{x}_t \) at diffusion time \( t \) into \( \bm{x}_{t-\Delta} \) at time \( t-\Delta \). 
Next, we use the update operator \(\mathcal{U}_\theta(\bm{x}_t) \defeq \) Eq.~\eqref{eq:sgld_cont} if $\bm{x}_t$ is continuous or Eq.~\eqref{eq:discrete_reverse} if categorical.

We first establish a convergence criterion on the proximity to the optimum, showing that as diffusion progresses, the projected updates remain close to the highest-likelihood regions of the data distribution while respecting constraints.
\begin{theorem}[Convergence Proximity]
\label{thrm:conv}
If \(\log p_\text{data}(\bm{x}_0) \) is convex, then there exists a minimum iteration \( \bar{t} \) such that, for all \( t \leq \bar{t} \), the following inequality holds:
\(
\|\mathcal{U}_\theta(\bm{x}_t) - \Phi\|_2 \leq \|\rho - \Phi\|_2,
\)
where \( \rho \) is the closest point to \( \Phi \), the global optimum  of \(\log p_\text{data}(\bm{x}_0)\), which can be reached via a single gradient step from any point in \( \mathbf{C} \).
\end{theorem}
Next, we show that incorporating projections systematically reduces the cost of enforcing feasibility, making the projection steps increasingly efficient as sampling progresses.
\begin{theorem}[Error Reduction via Projection]
\label{eq:theorem-main}
Let \( \mathcal{P}_\mathbf{C} \) be the projection operator onto \( \mathbf{C} \). For all \( t \leq \bar{t} \), as defined by Theorem \ref{thrm:conv}. Then,
\[
\mathbb{E} \left[ \text{Error}(\mathcal{U}_\theta(\bm{x}_t), \mathbf{C}) \right] \geq \mathbb{E} \left[ \text{Error}(\mathcal{U}_\theta(\mathcal{P}_\mathbf{C}(\bm{x}_t)), \mathbf{C}) \right],
\]
where \(\text{Error}(\cdot, \mathbf{C})\) quantifies the cost of projection.  
\end{theorem}

In essence, performing an update starting from a projected \(\bm{x}_t\) yields a sample \( \bm{x}_{t-\Delta} \) that is, in expectation, closer to the feasible set than an update without projection.
A direct consequence is:
\begin{corollary}[Convergence to Feasibility]
\label{cor:convergence}
For any arbitrarily small \(\xi>0\), there exists a time \(t\) such that after the update
\[
\mathrm{Error}\bigl(\mathcal{U}_\theta(\mathcal{P}_{\mathbf{C}}(\bm{x}_t)),\mathbf{C}\bigr) \; \le \; \xi.
\]
\end{corollary}
This result leverages the fact that the step size \(\gamma_t\) strictly decreases as \(t\) decreases, and thus, both the gradient magnitude and noise diminish. Consequently, the projection error approaches zero, implying that the updates steer the sample toward the feasible subdistribution of \(p_{\text{data}}(\bm{x}_0)\).

Together, Theorem~\ref{eq:theorem-main} and Corollary~\ref{cor:convergence}  explain why integrating the projection steps into the reverse update \(\bm{x}_t \to \bm{x}_{t-\Delta}\) produces samples that conform closely to the imposed constraints. 

\paragraph{Feasibility Guarantees.}
For an arbitrary density function,
NSD provides feasibility guarantees for convex constraint sets. 
This assurance is critical in applications of interest to this paper, including the material design explored in Section \ref{subsec:morph_prop} and physics-based simulations (Appendix \ref{appendix:phys}). 

\vspace{-12pt}
\section{Experiments and Evaluation}
\label{section:exp}

The evaluation of the symbolic diffusion approach focuses on three primary tasks 
designed to stress-test compliance with challenging constraints, with 
focus on \emph{safety}, \emph{data scarcity robustness}, 
and \emph{out-of-domain generalization}. 
In all cases, we compare our method (NSD) against 
state-of-the-art baseline diffusion models and relevant ablations, as assessed by domain-specific qualitative metrics (i.e., path length for motion planning and FID scores for image generation) and frequency of constraint violations. 
Complimenting any domain specific baselines, the evaluation also includes, where applicable, a \emph{conditional diffusion model (Cond)}, where constraints are applied 
as conditioning variables of the models and a \emph{post-hoc correction ($\text{Post}^+$)} approach (projecting the final output only) to illustrate the importance of integrating constraints during sampling. We use identical neural network architectures and training procedures for the diffusion model across methods; thus, differences in performance can be attributed to constraint implementation rather than model capabilities. 
Due to space constraints, we fully elaborate all domain specifications in Appendix \ref{appendix:experiments}.
To demonstrate the broad applicability of NSD, the experimental settings showcase its capability in:
\begin{enumerate}[label=\textbf{\arabic*.}, 
                  leftmargin=*, 
                  itemsep=0pt, 
                  parsep=0pt, 
                  topsep=0pt]
    \item Enabling \emph{safe, non-toxic} molecular generation and \emph{out-of-domain} discovery (\S\ref{subsec:safe_mol}).
    \item Handling \emph{safety-critical} settings and \emph{highly non-convex constraints} for motion planning (\S\ref{subsec:mrmp}).
    \item Facilitating microstructure design in \emph{data scarce} settings for \emph{out-of-domain} discovery (\S\ref{subsec:morph_prop}).
\end{enumerate}
In addition, we test the ability of NSD 
to generate ODE-governed videos for \emph{out-of-distribution} tasks (Appendix \ref{appendix:phys}), to prevent \emph{harmful} text generation for natural language modeling (Appendix \ref{appendix:safe_text}), and to constrain supplementary \emph{out-of-domain} molecule generation properties (Appendix \ref{appendix:molecule}).

\begin{figure}[t!]
\centering
\begin{minipage}[t]{0.57\linewidth}
  \vspace{0pt}  

    \centering
    \includegraphics[width=0.99\linewidth]{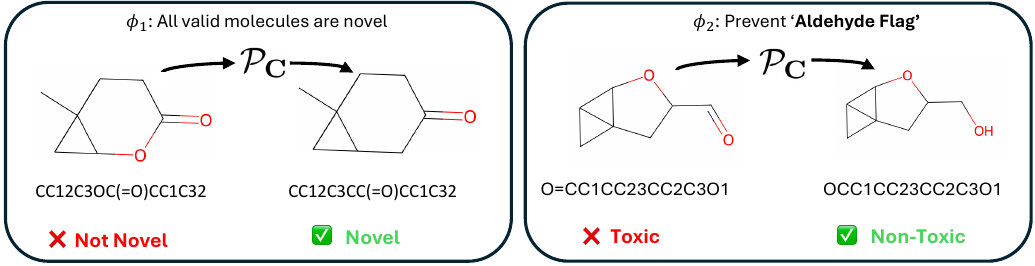}
  
\end{minipage}%
\hfill
\begin{minipage}[t]{0.42\linewidth}
  \vspace{1pt}  
  \centering
  \renewcommand{\arraystretch}{1.35}
  \resizebox{\linewidth}{!}{
  \begin{tabular}{|l|l|cc|cc|}
  \hline
  \multirow{7}{*}{\rotatebox{90}{\footnotesize \textbf{Molecule Generation}}}
    & \multirow{2}{*}{\textbf{Model}} 
    & \multirow{2}{*}{\textbf{Novel}} 
    & \multirow{2}{*}{\begin{tabular}{@{}c@{}} \textbf{Novel \&}\\ \textbf{ Non-Toxic} \end{tabular}} 
    & \multicolumn{2}{c|}{\textbf{Viol (\%)}} \\
    & & & 
    & \(\phi_1{: \text{novelty}}\)
    & \(\phi_{2:6}{: \text{non-toxic}}\) \\
  \cline{2-6}
  & AR 
    & \(\displaystyle 10.3 \pm 2.3\)  
    & \(\displaystyle 5.3 \pm 1.4\)  
    & \(\displaystyle 99.0 \pm 0.2\)  
    & \(\displaystyle 40.2 \pm 10.9\)  \\
  & MDLM
    &  \(\displaystyle 260.7 \pm 16.4\)  
    & \(\displaystyle 108.0 \pm 9.7\)   
    & \(\displaystyle 	53.9 \pm 3.1\)   
    & \(\displaystyle 	35.3 \pm 0.5\) \\
  & UDLM
    & \(\displaystyle 279.7 \pm 22.7\) 
    & \(\displaystyle 132.3 \pm 3.7\)
    & \(\displaystyle 70.8 \pm 2.4\) 
    & \(\displaystyle 38.1 \pm 3.3\) \\
  & $\textbf{NSD}_{\text{ BRENK}}$ 
    & \(\displaystyle 451.7 \pm 19.5\)
    & \(\displaystyle 392.0 \pm 16.7\)
    & \(\displaystyle  51.2 \pm 2.0\)
    & \textbf{0.0 \(\pm\) 0.0} \\
  & $\textbf{NSD}_{\text{BRENK + Novel}}$ 
    & \(\displaystyle \mathbf{533.3 \pm 8.7}\)
    & \(\displaystyle \mathbf{474.3 \pm 5.7}\)
    & \(\displaystyle \mathbf{1.4 \pm 0.3}\)
    & \textbf{0.0 \(\pm\) 0.0} \\
  \hline
  \end{tabular}
  }
\end{minipage}
\caption{Results for Molecule Generation experiments constrained to be novel and non-toxic. On the left, we provide examples of the projection operators; on the right, the table outlines specific results.}
\label{tab:sentence_molecules}
\end{figure}

\subsection{Molecule Generation for Drug Discovery (Safety and Domain Generalization)}
\label{subsec:safe_mol}
In drug discovery, ensuring the \emph{chemical safety and quality} of output molecules is critical. This experiment generates molecules in SMILES format \citep{weininger1988smiles} using a uniform discrete diffusion model finetuned on the QM9 dataset \citep{qm91, qm92}. For this task, NSD is compared with MDLM \citep{sahoo2024simple} and UDLM \citep{schiff2024simple}, the current state-of-the-art discrete diffusion models, and an autoregressive (AR) baseline with identical architecture and size to our diffusion model backbone. 

The experiment enforces two key constraints: a novelty constraint (\(\phi_1\)) that ensures generated molecules do not appear in the training set, and five BRENK substructure filters (\(\phi_{2:6}\)) that identify undesirable molecular fragments (e.g., aldehydes, three-membered heterocycles) linked to toxicity and the absence of drug-like characteristics. Critically, molecules flagged by BRENK often exhibit toxicity, reactivity, or other liabilities making them unsuitable for drug discovery \citep{brenk2008lessons}. Thus, for this study, 
we define `non-toxic' molecules as those passing all five of the chosen BRENK filters (Appendix \ref{appendix:molecule}).
An illustration of the NSD correction mechanism employed to project the generated molecules is presented in Figure~\ref{tab:sentence_molecules} (left).

Together, these constraints serve two purposes: \textbf{(1) Out-of-Distribution Generation:} the novelty constraint promotes \emph{out-of-distribution generation}, which is essential for discovering new chemical compounds, and \textbf{(2) Safety-Critical Outputs:} the BRENK filters \emph{ensure the sampled molecules are safe}, thereby improving their likelihood of success in downstream drug-development pipelines.

Figure~\ref{tab:sentence_molecules} (right) reports the number of novel and non-toxic molecules generated, along with constraint violations (expressed as the percentage of generations that do not conform to the imposed requirements). 
While the diffuison baselines report substantial improvements with respect to the AR model, they frequently violate constraints. In contrast, 
NSD achieves \emph{perfect adherence to safety constraints}, while also increasing the frequency of molecule generations that are novel, valid, and non-toxic \textit{by over 3.5\(\times\)},
a remarkable improvement over the current state-of-the-art. 
Additionally, as detailed in Appendix \ref{appendix:molecule}, we provide an evaluation of settings where NSD generations comply with strict thresholds on synthetic accessibility, the ease with which the generated molecules can be synthesized, further improving the practical utility.
 

\begin{figure}[t!]
\begin{minipage}{0.45\linewidth}
    \centering
    \includegraphics[width=0.8\columnwidth]{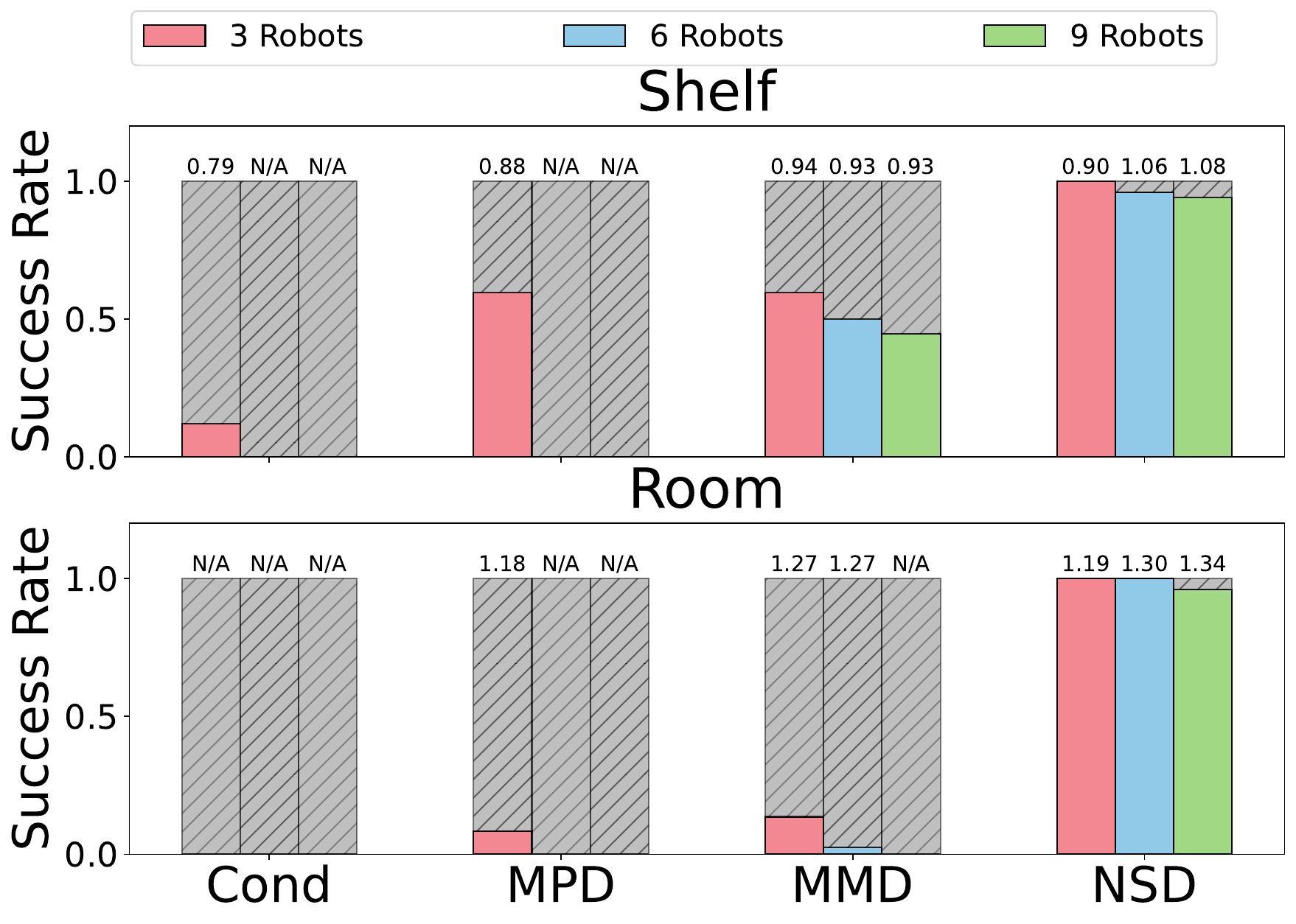}
    \label{fig:Results of Practical Maps}
\end{minipage}
\hfill
\begin{minipage}{0.52\linewidth}
    \centering
    \vspace{-15pt}
    \subfigure[Shelf Maps.]
    {
        \includegraphics[width=0.45\columnwidth]{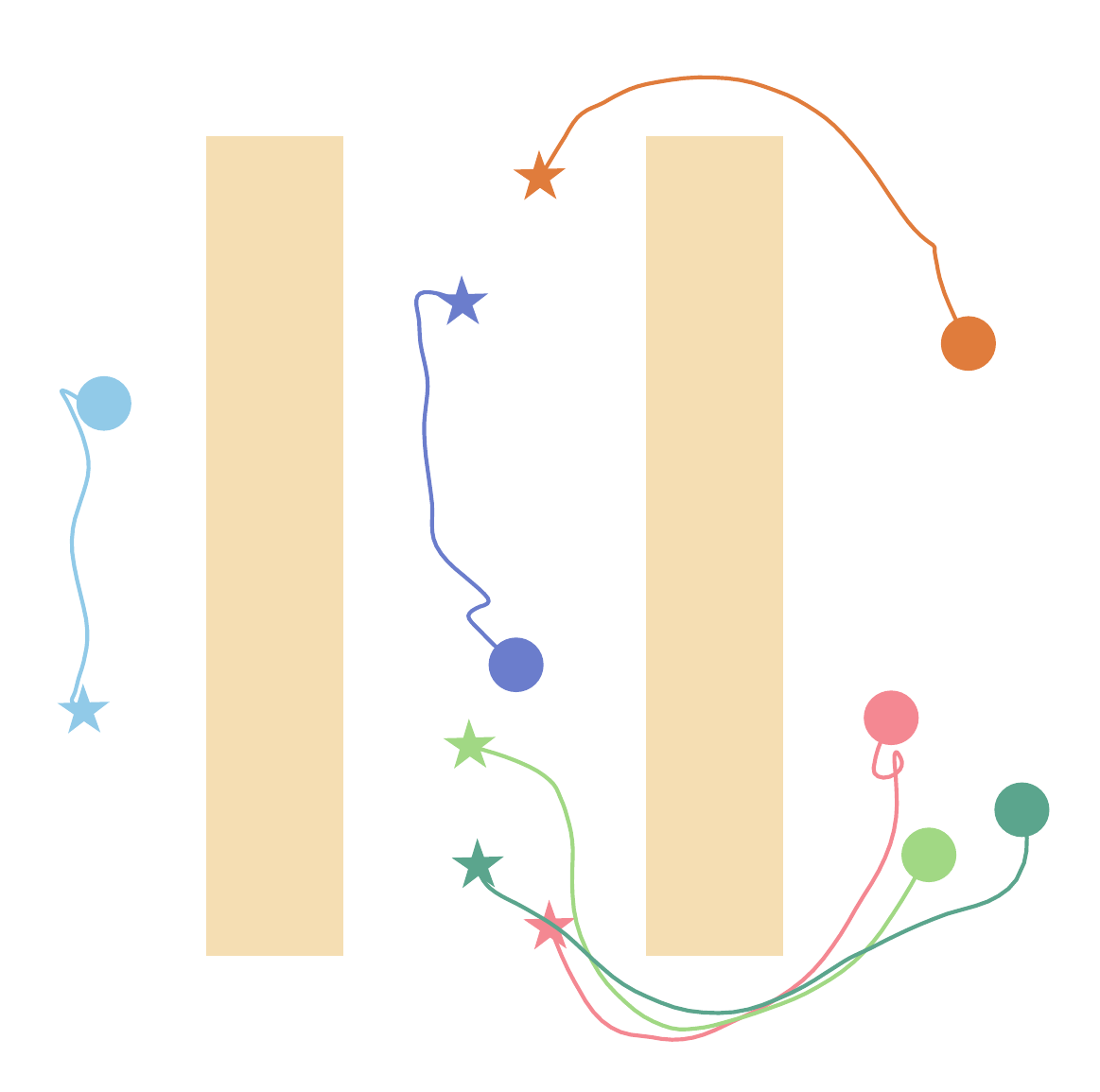}
        \label{fig:Trajectories for Shelf Maps}
    }
    \subfigure[Room Maps.]
    {
        \includegraphics[width=0.45\columnwidth]{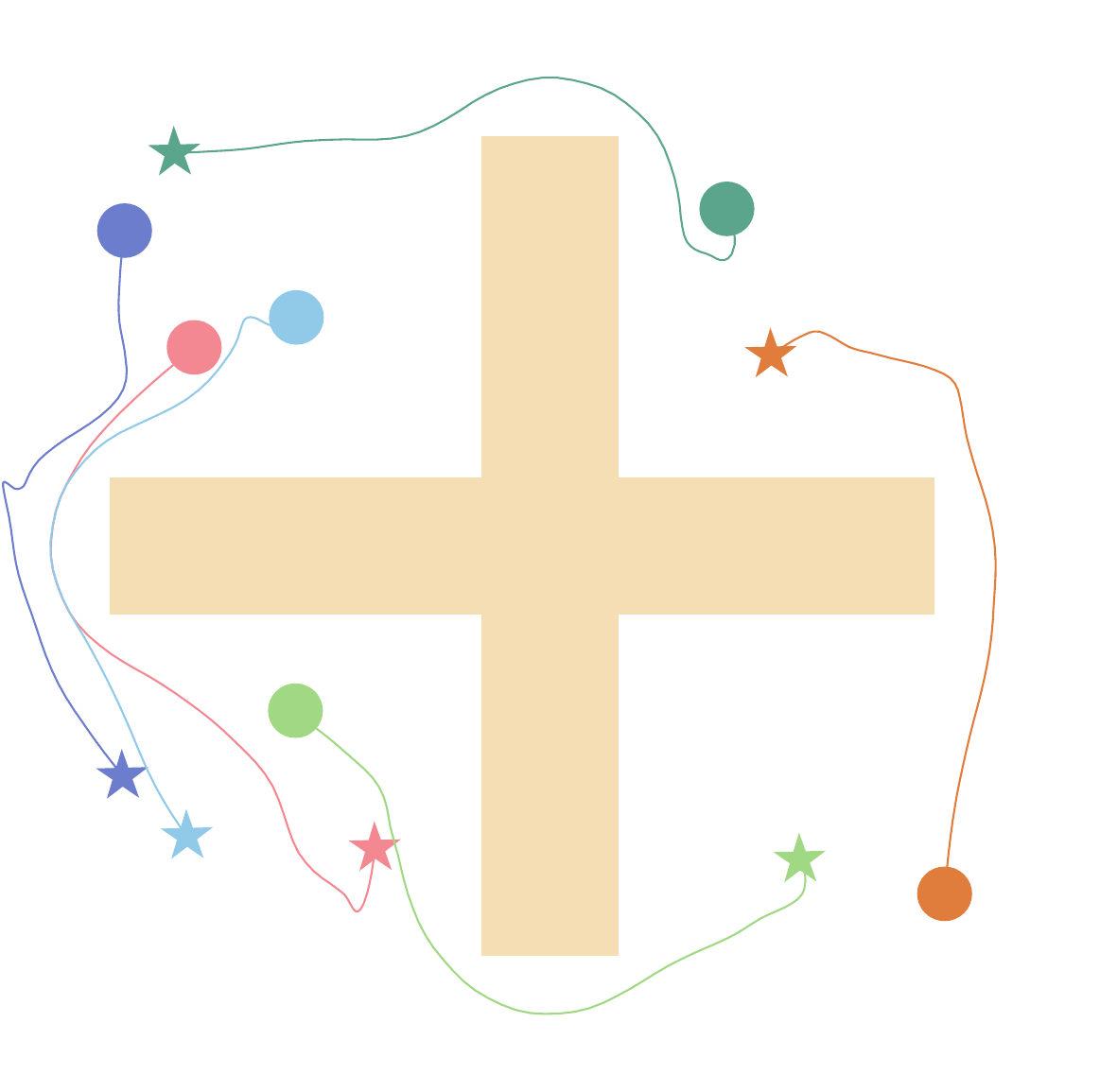}
        \label{fig:Trajectories for Room Maps}
    }
    \label{fig:Trajectories of Practical Maps}
\end{minipage}
\caption{Evaluation on practical maps with three different numbers of robots. On the left, we assess failure rates (depicted by the gray regions of the bars) and average path length (values on top of the bars). On the right, we provide visualizations of the practical maps tested on.}
\label{fig:motion_planning}
\end{figure}

\vspace{-6pt}
\subsection{Motion Planning for Autonomous Agents (Safety)}
\label{subsec:mrmp}
Next, we examine how NSD enforces collision avoidance in autonomous multi-agents settings using a \emph{continuous} diffusion model. 
Two main challenges arise: 
\textbf{(1) Safety-Critical Outputs:}
In real-world deployments, robots must avoid restricted or hazardous areas to ensure safe navigation in cluttered or dynamic environments. 
\textbf{(2) Highly Non-Convex Constraints:}
Furthermore, the underlying problem is characterized by a \emph{large number of non-convex} and temporal constraints, rendering the problem extremely challenging (more details in Appendix \ref{appendix:motion}).

We compare NSD to a conditional baseline and current state-of-the-art methods for multi-agent pathfinding: 
(1) Motion Planning Diffuion (MPD) originally designed for single-robot motion planning~\citep{carvalho2023motion}, and adapted here to multi-agent tasks for comparison and
(2) Multi-robot Motion Planning (MMD), a recent method that integrates diffusion models with classical search techniques \citep{shaoul2024multi}.
Figure \ref{fig:motion_planning} (left) highlights the results on practical maps that feature multiple rooms connected by doors or constrained pathways for robot navigation. These scenarios require robots to not only avoid collisions (\(\phi_{1}\)) but also coordinate globally to find feasible routes through shared spaces (\(\phi_{2}\)).
NSD sets a new \emph{state-of-the-art} in feasibility and scalability. 
In Figure~\ref{fig:Trajectories for Shelf Maps}, MPD and MMD achieve 60\% success with three robots but degrade significantly with more agents, with MMD dropping to 45\% for nine robots. In contrast, NSD maintains high success rates: 100\% for three robots, 96\% for six, and 93\% for nine. On room maps (Figure~\ref{fig:Trajectories for Room Maps}), NSD achieves perfect success rates for up to six robots and over 95\% for nine robots, whereas MPD and MMD fail entirely as complexity increases. This trend persists across other environments (see Appendix~\ref{appendix:motion}).
This is remarkable, as NSD can enable effective coordination among multiple robots in shared, constrained spaces, ensuring collision-free, kinematically feasible trajectories.

\vspace{-6pt}
\subsection{Morphometric Property Specification (Data Scarcity and Domain Generalization)}
\label{subsec:morph_prop}

\begin{wrapfigure}[14]{r}{0.5\linewidth}
\centering
\vspace{-18pt}

\begin{minipage}{0.5\textwidth}
  \ra{0.25}
  \setlength{\tabcolsep}{1pt}
  \centering
  \renewcommand{\arraystretch}{0.42}
  \begin{tabular}{c ccccc}
    \toprule
    \multirow{2}{*}{\footnotesize{Ground}} & \multirow{2}{*}{\footnotesize{P(\%)~~}} 
    & \multicolumn{4}{c}{\footnotesize{\textbf{Generative Methods}}}\\[2pt]
    \cline{3-6}\\[-2pt]
     & &  \footnotesize{\sl \bcol{NSD}} & \footnotesize{\sl {Cond}} & \footnotesize{\sl Post$^+$} & \footnotesize{\sl Cond$^+$} \\
    \midrule
    \includegraphics[width=0.16\columnwidth]{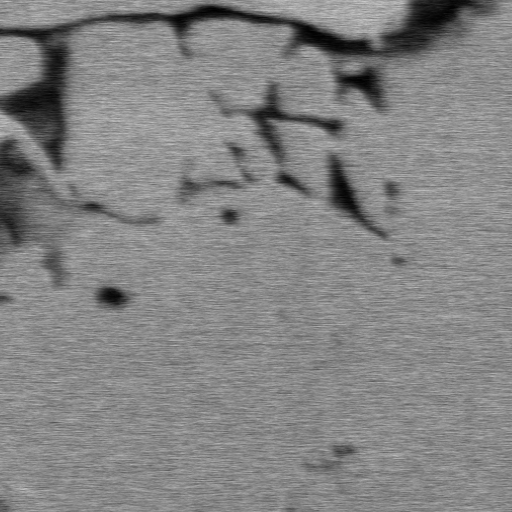} &
    {\footnotesize 10} &
    \includegraphics[width=.16\columnwidth]{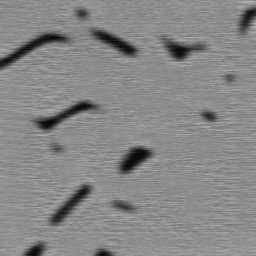} &
    \includegraphics[width=.16\columnwidth]{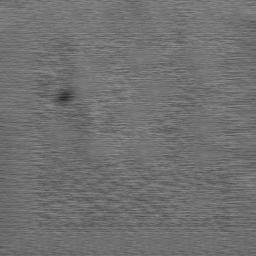} &
    \includegraphics[width=.16\columnwidth]{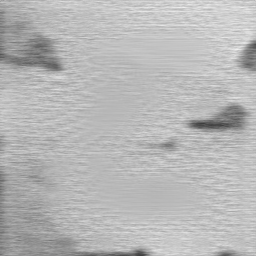} &
    \includegraphics[width=.16\columnwidth]{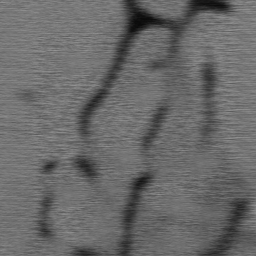} \\
    \includegraphics[width=0.16\columnwidth]{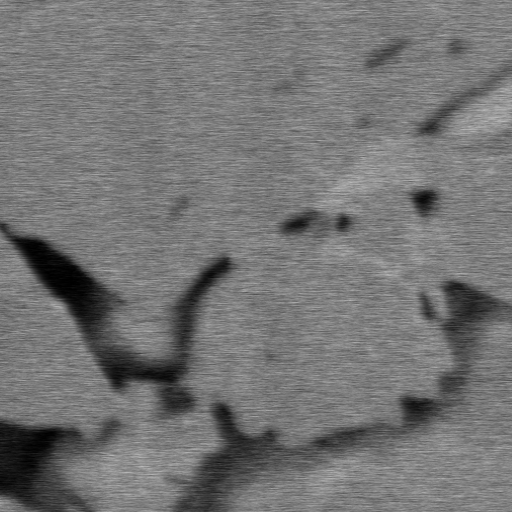} &
    {\footnotesize 30} &
    \includegraphics[width=.16\columnwidth]{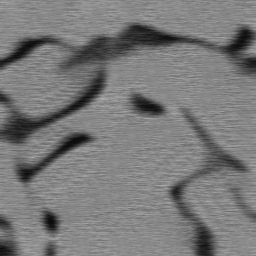} &
    \includegraphics[width=.16\columnwidth]{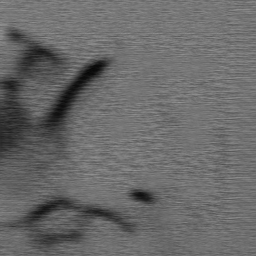} &
    \includegraphics[width=.16\columnwidth]{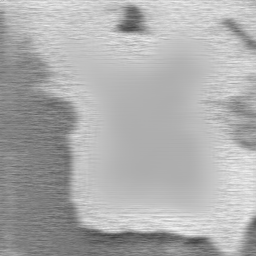} &
    \includegraphics[width=.16\columnwidth]{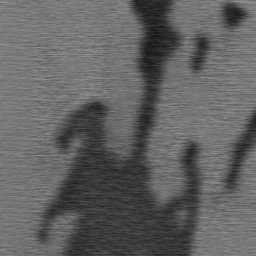} \\
    \includegraphics[width=0.16\columnwidth]{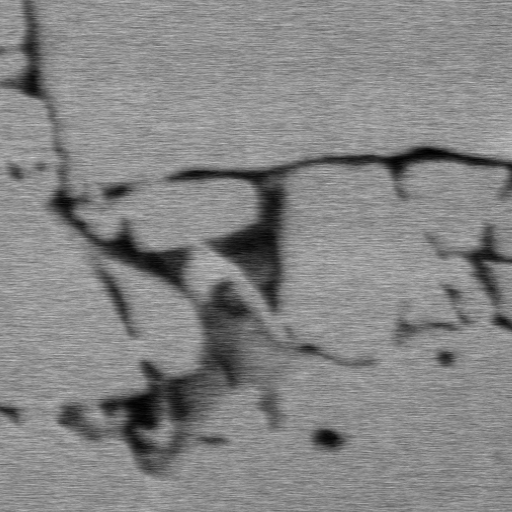} &
    {\footnotesize 50} &
    \includegraphics[width=.16\columnwidth]{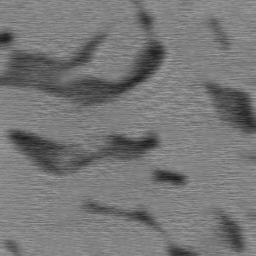} &
    \includegraphics[width=.16\columnwidth]{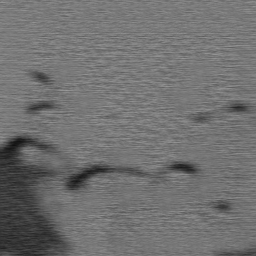} &
    \includegraphics[width=.16\columnwidth]{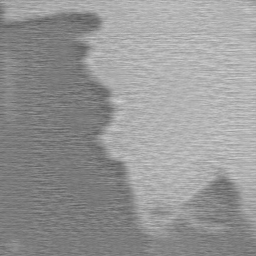} &
    \includegraphics[width=.16\columnwidth]{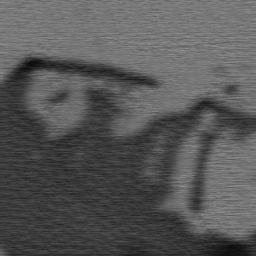} \\[2pt]
    \cline{3-6}\\
    \multicolumn{2}{r}{\textbf{ \scriptsize{FID:}}} & \tiny{\textbf{ 30.7 $\!\pm\!$ 6.8}} & \tiny{31.7 $\!\pm\!$ 15.6} & \tiny{41.7 $\!\pm\!$ 12.8} & \tiny{46.4 $\!\pm\!$ 10.7} \\
    \multicolumn{2}{r}{\textbf{\scriptsize{Viol. \(>\) 5\%:}}} & \tiny{\textbf{0.0}} & \tiny{94.2} & \tiny{\textbf{0.0}} & \tiny{\textbf{0.0}}\\
    \bottomrule
  \end{tabular}
\end{minipage}%
\hfill
\caption{Samples and results from the morphometric property specification experiments.}
\label{fig:micro}
\end{wrapfigure}

Finally, this experiment focuses on a microstructure design task. Here, achieving specific morphometric properties is crucial for expediting the discovery of structure-property linkages.
We consider an inverse-design problem in which a target porosity percentage, denoted by P(\%), is desired. Here, the porosity is a measure of the percentage of `damaged' pixels in the microstructure. This setting provides two particular challenges: \textbf{(1) Data Scarcity:} A key consideration in this context is the expense of generating training data, making data augmentation an important application of this problem.
Obtaining real material microstructure images is expensive and time-consuming, with limited control over attributes such as porosity, crystal sizes, and volume fraction, often requiring a trial-and-error approach.
Provided these costs, \emph{our training data regime is very low}; we subsample a single 3,000\(\times\) 3,000 pixel image to compose the dataset.
\textbf{(2) Out-of-Distribution Constraints} 
Given the low amounts of data available, often the \emph{desired porosity levels are unobserved in the training set}. 
Recall that NSD \textbf{guarantees} \emph{strict adherence to specified porosity constraints}.
Figure~\ref{fig:micro} illustrates the effectiveness of our method in generating microstructures with precise porosity levels as attempted by prior works employing conditional models \citep{chun2020deep}. 
This demonstrates that our approach not only provides the highest fidelity to the training distribution but also outperforms baselines in producing valid microstructures as assessed by domain-specific heuristic metrics (Figure~\ref{fig:morph_heuristic}, bottom and Appendix \ref{appendix:microstructure}).

\section{Conclusion}
This paper presented a novel framework that integrates symbolic optimization into the diffusion process, ensuring generative models can meet stringent physical, structural, or operational constraints. By enforcing these constraints continuously, rather than relying on post-processing approaches or soft guidance schemes, the proposed neuro-symbolic generative approach shows a unique ability to handle safety-critical tasks, cope with limited or skewed data, and generalize to settings beyond the original training distribution. 
Empirical evaluations across domains including toxic compound avoidance, motion planning, and inverse-design for material science illustrate this ability and provide a new state-of-the-art in utility and constraint adherence. 
As evidenced by this work, the capability to embed and process diverse symbolic knowledge and functional constraints into diffusion-based models paves the way for more trustworthy and reliable applications of generative AI in scientific, engineering, and industrial contexts.

\acks{
This research is partially supported by NSF grant RI-2334936 and NSF CAREER Award 2401285. The authors acknowledge Research Computing at the University of Virginia for providing computational resources that have contributed to the results reported within this paper. The views and conclusions of this work are those of the authors only.
}

\bibliography{bib}

\begin{thebibliography}{48}
\providecommand{\natexlab}[1]{#1}
\providecommand{\url}[1]{\texttt{#1}}
\expandafter\ifx\csname urlstyle\endcsname\relax
  \providecommand{\doi}[1]{doi: #1}\else
  \providecommand{\doi}{doi: \begingroup \urlstyle{rm}\Url}\fi

\bibitem[Ahmed et~al.(2022)Ahmed, Wang, Chang, and Van~den Broeck]{ahmed2022neuro}
Kareem Ahmed, Eric Wang, Kai-Wei Chang, and Guy Van~den Broeck.
\newblock Neuro-symbolic entropy regularization.
\newblock In \emph{Uncertainty in Artificial Intelligence}, pages 43--53. PMLR, 2022.

\bibitem[Betker et~al.(2023)Betker, Goh, Jing, Brooks, Wang, Li, Ouyang, Zhuang, Lee, Guo, et~al.]{betker2023improving}
James Betker, Gabriel Goh, Li~Jing, Tim Brooks, Jianfeng Wang, Linjie Li, Long Ouyang, Juntang Zhuang, Joyce Lee, Yufei Guo, et~al.
\newblock Improving image generation with better captions.
\newblock \emph{Computer Science. https://cdn. openai. com/papers/dall-e-3. pdf}, 2\penalty0 (3):\penalty0 8, 2023.

\bibitem[Boyd(2004)]{boyd2004convex}
Stephen Boyd.
\newblock Convex optimization.
\newblock \emph{Cambridge UP}, 2004.

\bibitem[Brenk et~al.(2008)Brenk, Schipani, James, Krasowski, Gilbert, Frearson, and Wyatt]{brenk2008lessons}
Ruth Brenk, Alessandro Schipani, Daniel James, Agata Krasowski, Ian~Hugh Gilbert, Julie Frearson, and Paul~Graham Wyatt.
\newblock Lessons learnt from assembling screening libraries for drug discovery for neglected diseases.
\newblock \emph{ChemMedChem: Chemistry Enabling Drug Discovery}, 3\penalty0 (3):\penalty0 435--444, 2008.

\bibitem[Broda et~al.(2002)Broda, d’Avila Garcez, and Gabbay]{broda2002neural}
Krysia~B Broda, A~d’Avila Garcez, and D~Gabbay.
\newblock Neural-symbolic learning system: foundations and applications, 2002.

\bibitem[Carvalho et~al.(2023)Carvalho, Le, Baierl, Koert, and Peters]{carvalho2023motion}
Joao Carvalho, An~T Le, Mark Baierl, Dorothea Koert, and Jan Peters.
\newblock Motion planning diffusion: Learning and planning of robot motions with diffusion models.
\newblock In \emph{2023 IEEE/RSJ International Conference on Intelligent Robots and Systems (IROS)}, pages 1916--1923. IEEE, 2023.

\bibitem[Christopher et~al.(2025)Christopher, Baek, and Fioretto]{christopher2024constrained}
Jacob~K Christopher, Stephen Baek, and Nando Fioretto.
\newblock Constrained synthesis with projected diffusion models.
\newblock \emph{Advances in Neural Information Processing Systems}, 37:\penalty0 89307--89333, 2025.

\bibitem[Chun et~al.(2020)Chun, Roy, Nguyen, Choi, Udaykumar, and Baek]{chun2020deep}
Sehyun Chun, Sidhartha Roy, Yen~Thi Nguyen, Joseph~B Choi, HS~Udaykumar, and Stephen~S Baek.
\newblock Deep learning for synthetic microstructure generation in a materials-by-design framework for heterogeneous energetic materials.
\newblock \emph{Scientific reports}, 10\penalty0 (1):\penalty0 13307, 2020.

\bibitem[Dathathri et~al.(2019)Dathathri, Madotto, Lan, Hung, Frank, Molino, Yosinski, and Liu]{dathathri2019plug}
Sumanth Dathathri, Andrea Madotto, Janice Lan, Jane Hung, Eric Frank, Piero Molino, Jason Yosinski, and Rosanne Liu.
\newblock Plug and play language models: A simple approach to controlled text generation.
\newblock \emph{arXiv preprint arXiv:1912.02164}, 2019.

\bibitem[Ertl and Schuffenhauer(2009)]{ertl2009estimation}
Peter Ertl and Ansgar Schuffenhauer.
\newblock Estimation of synthetic accessibility score of drug-like molecules based on molecular complexity and fragment contributions.
\newblock \emph{Journal of cheminformatics}, 1:\penalty0 1--11, 2009.

\bibitem[Fioretto et~al.(2020)Fioretto, {Van Hentenryck}, Mak, Tran, Baldo, and Lombardi]{FvHMTBL:ecml20}
Ferdinando Fioretto, Pascal {Van Hentenryck}, Terrence W.~K. Mak, Cuong Tran, Federico Baldo, and Michele Lombardi.
\newblock Lagrangian duality for constrained deep learning.
\newblock In \emph{European Conference on Machine Learning}, volume 12461 of \emph{Lecture Notes in Computer Science}, pages 118--135. Springer, 2020.
\newblock \doi{10.1007/978-3-030-67670-4\_8}.
\newblock URL \url{https://doi.org/10.1007/978-3-030-67670-4\_8}.

\bibitem[Gehman et~al.(2020)Gehman, Gururangan, Sap, Choi, and Smith]{gehman2020realtoxicityprompts}
Samuel Gehman, Suchin Gururangan, Maarten Sap, Yejin Choi, and Noah~A Smith.
\newblock Realtoxicityprompts: Evaluating neural toxic degeneration in language models.
\newblock \emph{arXiv preprint arXiv:2009.11462}, 2020.

\bibitem[Giannone et~al.(2023)Giannone, Srivastava, Winther, and Ahmed]{giannone2023aligning}
Giorgio Giannone, Akash Srivastava, Ole Winther, and Faez Ahmed.
\newblock Aligning optimization trajectories with diffusion models for constrained design generation.
\newblock \emph{arXiv preprint arXiv:2305.18470}, 2023.

\bibitem[Guo et~al.(2024)Guo, Yuan, Yang, Chen, and Wang]{guo2024gradient}
Yingqing Guo, Hui Yuan, Yukang Yang, Minshuo Chen, and Mengdi Wang.
\newblock Gradient guidance for diffusion models: An optimization perspective.
\newblock \emph{arXiv preprint arXiv:2404.14743}, 2024.

\bibitem[Ho and Salimans(2022)]{ho2022classifier}
Jonathan Ho and Tim Salimans.
\newblock Classifier-free diffusion guidance.
\newblock \emph{arXiv preprint arXiv:2207.12598}, 2022.

\bibitem[Ho et~al.(2020)Ho, Jain, and Abbeel]{ho2020denoising}
Jonathan Ho, Ajay Jain, and Pieter Abbeel.
\newblock Denoising diffusion probabilistic models.
\newblock \emph{Advances in neural information processing systems}, 33:\penalty0 6840--6851, 2020.

\bibitem[Jang et~al.(2016)Jang, Gu, and Poole]{jang2017categorical}
Eric Jang, Shixiang Gu, and Ben Poole.
\newblock Categorical reparameterization with gumbel-softmax.
\newblock \emph{arXiv preprint arXiv:1611.01144}, 2016.

\bibitem[Landrum et~al.(2025)Landrum, Tosco, Kelley, Rodriguez, Cosgrove, Vianello, sriniker, Gedeck, Jones, NadineSchneider, Kawashima, Nealschneider, Dalke, Swain, Cole, Turk, Savelev, tadhurst cdd, Vaucher, Wójcikowski, Take, Scalfani, Walker, Ujihara, Probst, Lehtivarjo, Faara, guillaume godin, Pahl, and Monat]{greg_landrum_2025_14779836}
Greg Landrum, Paolo Tosco, Brian Kelley, Ricardo Rodriguez, David Cosgrove, Riccardo Vianello, sriniker, Peter Gedeck, Gareth Jones, NadineSchneider, Eisuke Kawashima, Dan Nealschneider, Andrew Dalke, Matt Swain, Brian Cole, Samo Turk, Aleksandr Savelev, tadhurst cdd, Alain Vaucher, Maciej Wójcikowski, Ichiru Take, Vincent~F. Scalfani, Rachel Walker, Kazuya Ujihara, Daniel Probst, Juuso Lehtivarjo, Hussein Faara, guillaume godin, Axel Pahl, and Jeremy Monat.
\newblock rdkit/rdkit: 2024\_09\_5 (q3 2024) release, January 2025.
\newblock URL \url{https://doi.org/10.5281/zenodo.14779836}.

\bibitem[Liu et~al.(2024)Liu, Zhang, Li, Yan, Gao, Chen, Yuan, Huang, Sun, Gao, He, and Sun]{liu2024sorareviewbackgroundtechnology}
Yixin Liu, Kai Zhang, Yuan Li, Zhiling Yan, Chujie Gao, Ruoxi Chen, Zhengqing Yuan, Yue Huang, Hanchi Sun, Jianfeng Gao, Lifang He, and Lichao Sun.
\newblock Sora: A review on background, technology, limitations, and opportunities of large vision models, 2024.
\newblock URL \url{https://arxiv.org/abs/2402.17177}.

\bibitem[Lou et~al.(2024)Lou, Meng, and Ermon]{loudiscrete}
Aaron Lou, Chenlin Meng, and Stefano Ermon.
\newblock Discrete diffusion modeling by estimating the ratios of the data distribution.
\newblock In \emph{Forty-first International Conference on Machine Learning}, 2024.

\bibitem[Maz{\'e} and Ahmed(2023)]{maze2023diffusion}
Fran{\c{c}}ois Maz{\'e} and Faez Ahmed.
\newblock Diffusion models beat gans on topology optimization.
\newblock In \emph{Proceedings of the AAAI Conference on Artificial Intelligence (AAAI), Washington, DC}, 2023.

\bibitem[Meng et~al.(2022)Meng, Choi, Song, and Ermon]{meng2022concrete}
Chenlin Meng, Kristy Choi, Jiaming Song, and Stefano Ermon.
\newblock Concrete score matching: Generalized score matching for discrete data.
\newblock \emph{Advances in Neural Information Processing Systems}, 35:\penalty0 34532--34545, 2022.

\bibitem[Motamed et~al.(2025)Motamed, Culp, Swersky, Jaini, and Geirhos]{motamed2025generativevideomodelslearn}
Saman Motamed, Laura Culp, Kevin Swersky, Priyank Jaini, and Robert Geirhos.
\newblock Do generative video models learn physical principles from watching videos?, 2025.
\newblock URL \url{https://arxiv.org/abs/2501.09038}.

\bibitem[Nichol and Dhariwal(2021)]{nichol2021improved}
Alexander~Quinn Nichol and Prafulla Dhariwal.
\newblock Improved denoising diffusion probabilistic models.
\newblock In \emph{International conference on machine learning}, pages 8162--8171. PMLR, 2021.

\bibitem[Perez et~al.(2022)Perez, Huang, Song, Cai, Ring, Aslanides, Glaese, McAleese, and Irving]{perez2022red}
Ethan Perez, Saffron Huang, Francis Song, Trevor Cai, Roman Ring, John Aslanides, Amelia Glaese, Nat McAleese, and Geoffrey Irving.
\newblock Red teaming language models with language models.
\newblock \emph{arXiv preprint arXiv:2202.03286}, 2022.

\bibitem[Power et~al.(2023)Power, Soltani-Zarrin, Iba, and Berenson]{power2023sampling}
Thomas Power, Rana Soltani-Zarrin, Soshi Iba, and Dmitry Berenson.
\newblock Sampling constrained trajectories using composable diffusion models.
\newblock In \emph{IROS 2023 Workshop on Differentiable Probabilistic Robotics: Emerging Perspectives on Robot Learning}, 2023.

\bibitem[Ramakrishnan et~al.(2014)Ramakrishnan, Dral, Rupp, and von Lilienfeld]{qm92}
Raghunathan Ramakrishnan, Pavlo Dral, Matthias Rupp, and Anatole von Lilienfeld.
\newblock Quantum chemistry structures and properties of 134 kilo molecules.
\newblock \emph{Scientific Data}, 1, 08 2014.
\newblock \doi{10.1038/sdata.2014.22}.

\bibitem[Robey et~al.(2024)Robey, Ravichandran, Kumar, Hassani, and Pappas]{robey2024}
Alexander Robey, Zachary Ravichandran, Vijay Kumar, Hamed Hassani, and George~J. Pappas.
\newblock Jailbreaking llm-controlled robots, 2024.
\newblock URL \url{https://arxiv.org/abs/2410.13691}.

\bibitem[Rockafellar(2023)]{rockafellar2023convergence}
R~Tyrrell Rockafellar.
\newblock Convergence of augmented lagrangian methods in extensions beyond nonlinear programming.
\newblock \emph{Mathematical Programming}, 199\penalty0 (1):\penalty0 375--420, 2023.

\bibitem[Rombach et~al.(2022)Rombach, Blattmann, Lorenz, Esser, and Ommer]{rombach2022high}
Robin Rombach, Andreas Blattmann, Dominik Lorenz, Patrick Esser, and Bj{\"o}rn Ommer.
\newblock High-resolution image synthesis with latent diffusion models.
\newblock In \emph{Proceedings of the IEEE/CVF conference on computer vision and pattern recognition}, pages 10684--10695, 2022.

\bibitem[Ruddigkeit et~al.(2012)Ruddigkeit, Deursen, Blum, and Reymond]{qm91}
Lars Ruddigkeit, Ruud Deursen, Lorenz Blum, and Jean-Louis Reymond.
\newblock Enumeration of 166 billion organic small molecules in the chemical universe database gdb-17.
\newblock \emph{Journal of chemical information and modeling}, 52, 10 2012.
\newblock \doi{10.1021/ci300415d}.

\bibitem[Sahoo et~al.(2024)Sahoo, Arriola, Schiff, Gokaslan, Marroquin, Chiu, Rush, and Kuleshov]{sahoo2024simple}
Subham~Sekhar Sahoo, Marianne Arriola, Yair Schiff, Aaron Gokaslan, Edgar Marroquin, Justin~T Chiu, Alexander Rush, and Volodymyr Kuleshov.
\newblock Simple and effective masked diffusion language models.
\newblock \emph{arXiv preprint arXiv:2406.07524}, 2024.

\bibitem[Schiff et~al.(2024)Schiff, Sahoo, Phung, Wang, Boshar, Dalla-torre, de~Almeida, Rush, Pierrot, and Kuleshov]{schiff2024simple}
Yair Schiff, Subham~Sekhar Sahoo, Hao Phung, Guanghan Wang, Sam Boshar, Hugo Dalla-torre, Bernardo~P de~Almeida, Alexander Rush, Thomas Pierrot, and Volodymyr Kuleshov.
\newblock Simple guidance mechanisms for discrete diffusion models.
\newblock \emph{arXiv preprint arXiv:2412.10193}, 2024.

\bibitem[Shaoul et~al.(2024)Shaoul, Mishani, Vats, Li, and Likhachev]{shaoul2024multi}
Yorai Shaoul, Itamar Mishani, Shivam Vats, Jiaoyang Li, and Maxim Likhachev.
\newblock Multi-robot motion planning with diffusion models.
\newblock \emph{arXiv preprint arXiv:2410.03072}, 2024.

\bibitem[Sharifi et~al.(2023)Sharifi, Yildirim, and Fallah]{sharifi2023towards}
Iman Sharifi, Mustafa Yildirim, and Saber Fallah.
\newblock Towards safe autonomous driving policies using a neuro-symbolic deep reinforcement learning approach.
\newblock \emph{arXiv preprint arXiv:2307.01316}, 2023.

\bibitem[Shen et~al.(2025)Shen, Jiang, Yang, Wang, Han, and Li]{shen2025understanding}
Yifei Shen, Xinyang Jiang, Yifan Yang, Yezhen Wang, Dongqi Han, and Dongsheng Li.
\newblock Understanding and improving training-free loss-based diffusion guidance.
\newblock \emph{Advances in Neural Information Processing Systems}, 37:\penalty0 108974--109002, 2025.

\bibitem[Shi et~al.(2024)Shi, Han, Wang, Doucet, and Titsias]{shi2024simplified}
Jiaxin Shi, Kehang Han, Zhe Wang, Arnaud Doucet, and Michalis~K Titsias.
\newblock Simplified and generalized masked diffusion for discrete data.
\newblock \emph{arXiv preprint arXiv:2406.04329}, 2024.

\bibitem[Song and Ermon(2019)]{song2019generative}
Yang Song and Stefano Ermon.
\newblock Generative modeling by estimating gradients of the data distribution.
\newblock \emph{Advances in neural information processing systems}, 32, 2019.

\bibitem[Song et~al.(2020)Song, Sohl-Dickstein, Kingma, Kumar, Ermon, and Poole]{song2020score}
Yang Song, Jascha Sohl-Dickstein, Diederik~P Kingma, Abhishek Kumar, Stefano Ermon, and Ben Poole.
\newblock Score-based generative modeling through stochastic differential equations.
\newblock \emph{arXiv preprint arXiv:2011.13456}, 2020.

\bibitem[Voleti et~al.(2022)Voleti, Jolicoeur-Martineau, and Pal]{voleti2022mcvd}
Vikram Voleti, Alexia Jolicoeur-Martineau, and Chris Pal.
\newblock Mcvd-masked conditional video diffusion for prediction, generation, and interpolation.
\newblock \emph{Advances in Neural Information Processing Systems}, 35:\penalty0 23371--23385, 2022.

\bibitem[Wang et~al.(2023)Wang, Zheng, Ma, Du, Kim, Spielberg, Tenenbaum, Gan, and Rus]{wang2023diffusebot}
Tsun-Hsuan Wang, Juntian Zheng, Pingchuan Ma, Yilun Du, Byungchul Kim, Andrew Spielberg, Joshua Tenenbaum, Chuang Gan, and Daniela Rus.
\newblock Diffusebot: Breeding soft robots with physics-augmented generative diffusion models.
\newblock \emph{arXiv preprint arXiv:2311.17053}, 2023.

\bibitem[Watson et~al.(2023)Watson, Juergens, Bennett, Trippe, Yim, Eisenach, Ahern, Borst, Ragotte, Milles, et~al.]{watson2023novo}
Joseph~L Watson, David Juergens, Nathaniel~R Bennett, Brian~L Trippe, Jason Yim, Helen~E Eisenach, Woody Ahern, Andrew~J Borst, Robert~J Ragotte, Lukas~F Milles, et~al.
\newblock De novo design of protein structure and function with rfdiffusion.
\newblock \emph{Nature}, 620\penalty0 (7976):\penalty0 1089--1100, 2023.

\bibitem[Weininger(1988)]{weininger1988smiles}
David Weininger.
\newblock Smiles, a chemical language and information system. 1. introduction to methodology and encoding rules.
\newblock \emph{Journal of chemical information and computer sciences}, 28\penalty0 (1):\penalty0 31--36, 1988.

\bibitem[Wittmann(2024)]{wittmann2024exploring}
Maximilian Wittmann.
\newblock Exploring the effect of anthropomorphic design on trust in industrial robots: Insights from a metaverse cobot experiment.
\newblock In \emph{2024 21st International Conference on Ubiquitous Robots (UR)}, pages 118--124. IEEE, 2024.

\bibitem[Xu et~al.(2018)Xu, Chen, Zou, and Gu]{xu2018global}
Pan Xu, Jinghui Chen, Difan Zou, and Quanquan Gu.
\newblock Global convergence of langevin dynamics based algorithms for nonconvex optimization.
\newblock \emph{Advances in Neural Information Processing Systems}, 31, 2018.

\bibitem[Ye et~al.(2025)Ye, Lin, Han, Xu, Liu, Liang, Ma, Zou, and Ermon]{ye2025tfg}
Haotian Ye, Haowei Lin, Jiaqi Han, Minkai Xu, Sheng Liu, Yitao Liang, Jianzhu Ma, James~Y Zou, and Stefano Ermon.
\newblock Tfg: Unified training-free guidance for diffusion models.
\newblock \emph{Advances in Neural Information Processing Systems}, 37:\penalty0 22370--22417, 2025.

\bibitem[Yuan et~al.(2023)Yuan, Song, Iqbal, Vahdat, and Kautz]{yuan2023physdiff}
Ye~Yuan, Jiaming Song, Umar Iqbal, Arash Vahdat, and Jan Kautz.
\newblock Physdiff: Physics-guided human motion diffusion model.
\newblock In \emph{Proceedings of the IEEE/CVF International Conference on Computer Vision}, pages 16010--16021, 2023.

\bibitem[Zheng et~al.(2024)Zheng, Chen, Mao, Liu, Zhu, and Zhang]{zheng2024masked}
Kaiwen Zheng, Yongxin Chen, Hanzi Mao, Ming-Yu Liu, Jun Zhu, and Qinsheng Zhang.
\newblock Masked diffusion models are secretly time-agnostic masked models and exploit inaccurate categorical sampling.
\newblock \emph{arXiv preprint arXiv:2409.02908}, 2024.

\end{thebibliography}

\newpage
\appendix

\section{Extended Related Work}

\textbf{Neuro-symbolic frameworks.}
This paper's novel integration of symbolic constraints with generative models builds on foundational work in hybrid AI systems, blending the pattern recognition of neural networks with symbolic reasoning’s structured constraints. Early approaches like cooperative architectures \citep{broda2002neural} established iterative feedback loops between neural and symbolic components, as seen in autonomous driving systems where visual detectors refine predictions via spatial logic rules \citep{sharifi2023towards}. Parallel efforts in compiled architectures embedded symbolic operations directly into neural activations, enabling dynamic constraint enforcement in domains such as finance, where neurons encoded regulatory thresholds into credit scoring models \citep{ahmed2022neuro}.

\noindent\textbf{Training-free correction.}
An alternative approach to enforcing desired properties in diffusion models is through training-free correction via gradient‐based guidance. Inspired by methods such as Plug and Play Language Models (PPLM) \citep{dathathri2019plug}, these techniques compute gradients from an external objective or constraint function at sampling time. Rather than relying on additional classifier training or extensive data labeling, the method directly adjusts the score estimates during the sampling process. Specifically, a loss function encoding the desired property is defined over the generated sample, and its gradient with respect to the sample is computed.
Unlike model conditioning, which augments the score with a fixed conditioning signal, training-free correction dynamically refines the generation by continuously monitoring and correcting deviations from the target behavior \citep{guo2024gradient, shen2025understanding}.
Such methods provide an alternative to existing conditioning approaches, but generally report worse performance than conditioning methods, due to inaccuracies in their gradients when the sample is at higher noise levels \citep{ye2025tfg}.

\section{Augmented Lagrangian Method}
\label{appendix:alm}

Since \(\tilde{\phi}\) is typically nonlinear and hard to enforce directly, we adopt an augmented Lagrangian approach \citep{boyd2004convex}, which embeds the constraint \(\tilde{\phi}(\bm{y})\approx 0\) into a minimization objective with multipliers \(\lambda\) and a quadratic penalty \(\mu\). Let \(\mathcal{U}_\theta(\bm{x}_t)\) be the sample after applying the denoising step at time \(t\). 
We introduce a projected sample \(\bm{y}\) that we iteratively refine to reduce violations of \(\tilde{\phi}\) while remaining close to \(\mathcal{U}_\theta(\bm{x}_t)\) under \(D_{\mathrm{cost}}\). The augmented Lagrangian is:
\[
\mathcal{L}_{\text{ALM}}\bigl(\bm{y}, \lambda, \mu\bigr) 
= D_{\mathrm{cost}}\!\bigl(\bm{x}_t, \bm{y}\bigr)
+ \lambda \,\tilde{\phi}(\bm{y})
+ \tfrac{\mu}{2}\,\tilde{\phi}(\bm{y})^2.
\]

Minimizing \(\mathcal{L}_{\mathrm{ALM}}\) yields a lower-bound approximation to the original projection. Its Lagrangian dual solves:
\[
\arg\max_{\lambda,\mu}
\Bigl(\arg\min_{\bm{y}}
\mathcal{L}_{\text{ALM}}\bigl(\bm{y}, \lambda, \mu\bigr)\Bigr).
\]

We optimize iteratively, updating \(\bm{y}\) via gradient descent and adjusting \(\lambda\) and \(\mu\) as follows \cite{FvHMTBL:ecml20}:
\begin{subequations}
\label{eq:ld_update_rewrite}
\begin{align}
    \bm{y} &\leftarrow \bm{y} - \gamma \nabla_{\bm{y}} \mathcal{L}_{\text{ALM}}\bigl(\bm{y}, \lambda, \mu\bigr), \\
    \lambda &\leftarrow \lambda + \mu\,\tilde{\phi}(\bm{y}), \\
    \mu &\leftarrow \min(\alpha \mu, \mu_{\max}),
\end{align}
\end{subequations}
where \(\gamma\) is the gradient step size, \(\alpha>1\) increases \(\mu\) over iterations, and \(\mu_{\max}\) is an upper bound. This drives \(\bm{y}\) to satisfy \(\tilde{\phi}(\bm{y}) \approx 0\) while staying close to \(\mathcal{U}_\theta(\bm{x}_t)\). 
Noting that \(\tilde{\phi}\) may be computed using a surrogate network, this optimization can be further grounded by directly using \(\phi_{1:n}(\bm{y}) = 1\) as the termination condition; hence, assuming strong convergence properties (which are encouraged by the inclusion of the quadratic term \citep{rockafellar2023convergence}), the projected sample will strictly satisfy the symbolic constraints as assessed by the reasoning test.

\section{Discrete Sequence Relaxations}
\label{appendix:gumbel}

An important challenge to imposing gradient-based projections on discrete data sequences is providing a differentiable relaxation of our constraint satisfaction metric. This arises because we impose the constraint over the decoded version of the probability distributions, which is inherently discrete, making it not naturally differentiable. 
This poses a significant obstacle when one needs to backpropagate errors through operations that select discrete tokens or decisions. To overcome this limitation, we leverage a Gumbel-Softmax relaxation of the \(\arg\max\) operator, denoted as \( \psi \), which effectively bridges the gap between discrete and continuous representations.  

More specifically, given a probability vector of token $i$, \( \bm{x}_t^i \), where each component \( \bm{x}_t^i(v) \) represents the probability assigned to token \( v \) from a vocabulary of size \( V \), we approximate the hard, discrete decision of the \( \arg\max \) function by constructing a continuous, differentiable approximation:
\[
    \psi(\bm{x}_t^i)(v) = \frac{\exp\!\Bigl(\frac{\log \bm{x}_t^i(v) + g_v}{\text{T}_\text{sample}}\Bigr)}{\sum_{v'=1}^{V} \exp\!\Bigl(\frac{\log \bm{x}_t^i(v') + g_{v'}}{\text{T}_\text{sample}}\Bigr)}.
\]
Here, \( g_{v} \) is a random variable drawn independently from a Gumbel\((0,1)\) distribution for each token \( v \). The introduction of the Gumbel noise \( g_v \) perturbs the log-probabilities, thereby mimicking the stochasticity inherent in the discrete sampling process. The parameter \( \text{T}_\text{sample} > 0 \) serves as a temperature parameter that governs the degree of smoothness of the resulting distribution. Lower temperatures make the approximation sharper and more similar to the original \( \arg\max \) operator, while higher temperatures yield a smoother distribution that is more amenable to gradient-based optimization.

This relaxation not only facilitates the propagation of gradients through the projection step but also maintains a close approximation to the original discrete decision process. By incorporating the Gumbel-Softmax technique, we can integrate the \( \arg\max \) operation into our model in a way that is compatible with gradient descent, ultimately enabling the end-to-end training of models that require discrete token decisions without sacrificing the benefits of differentiability~\citep{jang2017categorical}.

\section{Score Matching for Discrete Diffusion}
\label{appendix:concrete}

Recall that in Section \ref{sec:prelim} we introduce the Euler discretized update step for Langevin dynamics:
\[
 \bm{x}_{t - \Delta} = \bm{x}_t + \gamma_t \nabla_{\bm{x}_t} \log p(\bm{x}_t) + \sqrt{2\gamma_t} \epsilon  
\]
This directly allows us to formulate the objective of the reverse process from the given update rule, as shown in Equation \eqref{eq:rev_opt}. This representation of the diffusion sampling procedure is fundamental to our theoretical analysis. While our discussion focuses on continuous diffusion models, as noted earlier, the framework can be naturally extended to discrete diffusion models as well. 

Particularly, we highlight that Langevin dynamics sampling algorithms used by continuous \emph{score-based} diffusion models, whether applied directly \citep{song2019generative} or through predictor-corrector frameworks \citep{song2020score}, can be utilized by discrete diffusion models.
Notably, score-based discrete diffusion models leverage a discrete generalization of the score function, referred to as the \emph{Concrete score} \citep{meng2022concrete}, to approximate the gradient of the probability density function $\log p_t(\bm{x}_t)$. As opposed to continuous score-based diffusion, where the gradient is directly applied to the representation of the sample (e.g., for image data the gradient will directly change pixel values), discrete models apply this gradient to the probability distributions which are sampled from to predict the final, discrete sequence. Despite this discrepancy, Concrete Score Matching provides an approach which mirrors continuous Score Matching in that the estimated gradients of the probability density function are used to guide the sample to high density regions of the target distribution.

As a final note, while many works do not explicitly adopt Concrete Score Matching as done by previous literature \citep{meng2022concrete,loudiscrete}, the score function is often still implicitly modeled by the denoiser. For example, \cite{sahoo2024simple} provide theoretical results demonstrating equivalence to a score-based modeling, supporting the extrapolation of our theoretical framework to models which employ simplified derivations of the Concrete Score Matching training objective.

\section{Extended Experimental Results}
\label{appendix:experiments}

\subsection{Molecule Generation for Drug Discovery (Safety and Domain Generalization)}
\label{appendix:molecule}

In this section further explain the setting for constrained molecular generation. We use UDLM \citep{schiff2024simple}  as our underlying diffusion model architecture for NSD for this application. The task is to generate representations of molecule structures using SMILES sequences \citep{weininger1988smiles}, human readable strings that can be directly mapped to molecule compounds. 
We begin with an overview of the domain-specific benchmarks used in our evaluation.
Then, we provide an extended version of Figure~\ref{tab:sentence_molecules} (right), where we detail the violations for each specific BRENK test that is corrected by our projection operator. We then discuss the violations, their corresponding symbolics tests, and projection operators in detail. Finally, we introduce and explain the setting and results for constraining the synthetic accessibility of the molecules generated. 

\paragraph{Additional benchmarks.}
To supplement our evaluation, we compare to several domain specific approaches:
\begin{enumerate}[leftmargin=*, parsep=0pt, itemsep=0pt, topsep=0pt]
    \item \textbf{Autoregressive LLM (AR)}: An autoregressive transformer-based model, trained for molecule generations and sized to be comparable with the other diffusion-based architectures (100M parameters).
    \item \textbf{Conditional Masked Diffusion Model (MDLM)}: A conditional masked discrete diffusion model implementation from \cite{schiff2024simple} with guidance schemes in the subscript if applicable.
    \item \textbf{Conditional Uniform Diffusion Model (UDLM)}: A conditional uniform discrete diffusion model implementation from \cite{schiff2024simple} with guidance schemes in the subscript if applicable.
\end{enumerate}

\begin{table}[ht!]
\begin{minipage}[h]{\linewidth}
  \vspace{0pt}  
  \centering
  \renewcommand{\arraystretch}{1.235}
  \resizebox{\linewidth}{!}{
  \centering
  
  \begin{tabular}{|l|l|c|c|c|c|c|c|c|}
  \hline
  \multirow{8}{*}{\rotatebox{90}{\footnotesize \textbf{Molecule Generation}}}
    & \multirow{2}{*}{\textbf{Model}} 
    & \multirow{2}{*}{\begin{tabular}{@{}c@{}} \textbf{Novel \&}\\ \textbf{Non-Toxic} \end{tabular}} 
    & \multicolumn{6}{c|}{\textbf{Viol (\%)} } \\
    & & 
    & \(\phi_1{: \text{novelty}}\)
    & \(\phi_2{: \text{Aldehydes}}\)
    & \(\phi_3{: \text{Three-membered heterocycles}}\)
    & \(\phi_4{: \text{Imines}}\)
    & \(\phi_5{: \text{Triple bonds}}\)
    & \(\phi_6{: \text{Isolated alkene}}\) \\
  \cline{2-9}
  & AR 
    & \(\displaystyle 5.3 \pm 1.4\)  
    & \(\displaystyle 99.0 \pm 0.2\)  
    & \(\displaystyle 20.2 \pm 6.2\)
    & \(\displaystyle 9.3 \pm 7.6\)
    & \(\displaystyle 21.2 \pm 13.9\)
    & \(\displaystyle 10.9 \pm 2.6\)
    & \(\displaystyle 2.6 \pm 4.1\) \\
  & MDLM
    & \(\displaystyle 108.0 \pm 9.7\)   
    & \(\displaystyle 53.9 \pm 3.1\)   
    & \(\displaystyle 9.5 \pm 1.4\)
    & \(\displaystyle 22.2 \pm 2.1\)
    & \(\displaystyle 11.0 \pm 1.9\)
    & \(\displaystyle 9.0 \pm 1.5\)
    & \(\displaystyle 10.3 \pm 0.9\) \\
  & UDLM
    & \(\displaystyle 132.3 \pm 3.7\)
    & \(\displaystyle 70.8 \pm 2.4\) 
    & \(\displaystyle 11.3 \pm 1.4\)
    & \(\displaystyle 16.9 \pm 2.1\)
    & \(\displaystyle 10.9 \pm 2.6\)
    & \(\displaystyle 8.0 \pm 2.9\)
    & \(\displaystyle 11.1 \pm 1.6\) \\
  & \(\textbf{NSD}_{\text{ BRENK}}\) 
    & \(\displaystyle 392.0 \pm 16.7\)
    & \(\displaystyle 51.2 \pm 2.0\)
    & \textbf{0.0 \(\pm\) 0.0}
    & \textbf{0.0 \(\pm\) 0.0}
    & \textbf{0.0 \(\pm\) 0.0}
    & \textbf{0.0 \(\pm\) 0.0}
    & \textbf{0.0 \(\pm\) 0.0} \\
  & \(\textbf{NSD}_{\text{Novel + BRENK}}\) 
    & \(\displaystyle 474.3 \pm 5.7\)
    & \(\displaystyle 1.4 \pm 0.3\)
    & \textbf{0.0 \(\pm\) 0.0}
    & \textbf{0.0 \(\pm\) 0.0}
    & \textbf{0.0 \(\pm\) 0.0}
    & \textbf{0.0 \(\pm\) 0.0}
    & \textbf{0.0 \(\pm\) 0.0} \\
  \hline
  \end{tabular}
  }
\end{minipage}
\caption{Extended results from Figure~\ref{tab:sentence_molecules} (right).}
\label{fig:extended_results}
\end{table}

\paragraph{Symbolic test.}  
For the purpose of generating safer, higher quality, and novel molecules we implement a total of six symbolic tests each corresponding with its own correction; in practice, our projection operator composes these corrections to find the nearest feasible point \emph{at the intersection} of these constraints, \(\bm{x} \in \mathbf{C}\).

\begin{enumerate}[leftmargin=*, parsep=0pt, itemsep=0pt, topsep=0pt]
    \item \textbf{Novelty $\phi_{1}$}: As defined in \citep{schiff2024simple}, a generate molecule is considered valid if it can be parsed by the RDKit library \citep{greg_landrum_2025_14779836}, and a molecule is novel if its valid, unique, and not present in the QM9 dataset \citep{qm91, qm92}. The test function $\phi_{1}$ determines whether the current discretized molecular representation $\bm{x}^\star$ is within the training set $\mathcal{D}$, thus $\phi_1 \overset{\text{def}}{=} \bm{x}^\star \notin \mathcal{D}$. If $\phi_1$ is not satisfied, the corresponding projection operation uses a best-first search to systematically flip tokens in a sequence, based on minimal probability “flip costs” to generate new sequences not yet in the dataset. Once a novel sequence is found, it is finalized, added to the dataset, and the model’s token probabilities are adjusted to maintain high likelihood for the newly generated sequence.
    Specifically, we seek a novel sequence \(\bm{x}^\star \notin \mathcal{D}\) by minimally altering the top-ranked tokens and flip cost denoted as $D_{\mathrm{cost}}$.
We find 
\[
     \mathcal{P}_\mathbf{C}(\bm{x}) 
    \defeq 
    \underset{\bm{y}^\star \notin \mathcal{D}}{\mathrm{argmin}} 
    \; D_{\mathrm{cost}}(\bm{y}, \bm{x}),
\]
via a best-first search. 
Once a novel sequence is found, it is added to \(\mathcal{D}\), 
and the distribution is updated so that \(\bm{x}\) becomes the new top-ranked path, 
avoiding duplicates in future generations.

    \item \textbf{Substructure Violations $\phi_{2:6}$}: In order to generate safer and less toxic molecules we use the blackbox BRENK filter \citep{brenk2008lessons} provided by RDKit \citep{greg_landrum_2025_14779836} which offers various violation alerts which lead to a BRENK flag. While there are many of these potential substructure violations, we cover the five most frequent. For these violation we use RDKit to identify and flag these substructures. 
        Thus, we can define $B = \{\, \bm{x}^\star \mid \mathrm{BRENK_{Flag}}(\bm{x}^\star) = \mathrm{True}\},$ where \(B\) is the set of molecules that the BRENK filter flags. Now, we can define the tests as: $\phi_{2:6} \overset{\text{def}}{=}  \bm{x}^\star \notin B$ with each specific $\phi_i$ described below.

     \begin{enumerate}[label=(\alph*), leftmargin=*, itemsep=0pt, topsep=0pt]
         \item \textbf{Aldehydes $\phi_{2}$}: Aldehydes feature a carbonyl group (\texttt{C=O}) in which the carbon is also bonded to at least one hydrogen (i.e., \texttt{R--CHO}). In SMILES notation, this typically appears as \texttt{C=O} where the carbon atom carries a hydrogen. In drug discovery, aldehydes are considered undesirable owing to their high reactivity and potential toxicity.

To address flagged aldehydes, our method proceeds as follows:
\begin{enumerate}
    \item \emph{Transform}: Identify the aldehyde (\texttt{C=O} with a hydrogen on the carbon) and attempt to convert it into either an alcohol (\texttt{R--CH$_2$--OH}) or a methyl ketone (\texttt{R--C(=O)CH$_3$}).
    \item \emph{Fallback}: If neither transformation produces a valid molecule, directly modify the carbonyl bond (e.g., force \texttt{C=O} to \texttt{C--OH}) or remove the oxygen entirely, thus eliminating the problematic functionality.
\end{enumerate}
These operations yield a molecular sequence that no longer violates the aldehyde-related BRENK filter. 
        \item \textbf{Three-membered heterocycles $\phi_{3}$}: These are small ring systems composed of three atoms, at least one of which is a heteroatom (e.g., nitrogen, oxygen, etc.). Molecules containing such rings are considered undesirable due to their high ring strain, reactivity, and potential toxicity. Typically, they are detected by filters that look for three-member rings containing a heteroatom.

    To correct a flagged three-membered heterocycle, we use a stepwise approach: 
    \begin{enumerate}
        \item \emph{Ring expansion}: Insert a new carbon atom into one of the ring bonds, creating a larger, less-strained ring.
        \item \emph{Bond breaking}: If expansion fails to produce a valid, non-flagged molecule, break one of the ring bonds to open the ring.
        \item \emph{Complete removal}: If neither of the previous steps works, remove all bonds in the original three-membered ring entirely.
    \end{enumerate}
    After each step, we check for validation and against the BRENK filter. The result is a structure that no longer violates the “three-membered heterocycle” constraint.
        \item \textbf{Imines $\phi_{4}$}: An imine is a functional group containing a carbon–nitrogen double bond (\texttt{C=N}). These groups are often flagged due to potential instability and reactivity.

    The corresponding operation employs a two-stage procedure:
    \begin{enumerate}
        \item \emph{Initial fix}: Convert the double bond into a single bond and add a hydrogen atom to the nitrogen.
        \item \emph{Fallback fix}: If the first approach fails or yields an invalid structure, remove or break the \texttt{C=N} bond entirely so that no imine remains.
    \end{enumerate}
        \item \textbf{Triple bonds $\phi_{5}$}: Molecules containing triple bonds (denoted by \texttt{\#} in SMILES) can be flagged due to concerns about reactivity, metabolic stability, or synthetic difficulty. To address such cases, we apply a simple transformation that replaces the triple bond character (\texttt{\#}) with either a double bond (\texttt{=}) or a single bond (\texttt{-}), thus reducing the likelihood of reactivity or instability. 
        \item \textbf{Isolated alkene $\phi_{6}$}: Alkenes, represented by (\texttt{=}) can be flagged if they appear in undesired or isolated positions that may lead to reactivity or instability issues. To address this, when flagged, our method replaces the double bond character (\texttt{=}) with a single bond (\texttt{-}), effectively saturating the alkene. This ensures that the final structure does not violate the isolated-alkene BRENK constraint.
    \end{enumerate}
\end{enumerate}

Next, to supplement the evaluations provided in the main text, we provide an additional setting where we constrain synthetic accessibility (SA) of the generated molecules below a strict threshold.
For this setting, we consider that many of the generated molecules, while potentially novel and valid generations, cannot be directly synthesized due to the complexity of these compounds \citep{ertl2009estimation}. Hence, we impose a constraint on the permissible synthetic accessibility of these outputs and compare to a series of conditional models (Table \ref{fig:sa}).
Notably, our model yields a 0\% violation rate, despite generating a competitive number of valid molecules and exhibiting the highest drug-likeness scores (referred to in the table as QED).
These results demonstrate how the inclusion of constraint projection operators ensure that generated molecules not only scores well in property optimization but also adhere to synthetic feasibility requirements as determined by an independent, external standard.

\begin{table}[!ht]
\centering
\begin{minipage}[t]{0.9\linewidth}
\vspace{0pt} 
\centering
\resizebox{\textwidth}{!}{
\begin{tabular}{|l|l| c c c |cccc|}
         \cline{1-9}
         \multirow{14}{*}{\rotatebox{90}{\footnotesize \textbf{Molecules (Synthetic Accessibility)}}}
         & \multirow{2}{*}{\textbf{Model}} 
         & \textbf{Valid} & \textbf{Novel} & \textbf{QED} & \multicolumn{4}{c|}{\textbf{Viol (\%)}}  \\
         &  & $[\uparrow]$ & $[\uparrow]$ & $[\uparrow]$ & \(\tau=3.0\) & \(\tau=3.5\) & \(\tau=4.0\) & \(\tau=4.5\) \\
         \cline{2-9}
         & $\text{AR}$                               & 1023 & 11  & 0.46 & 91.6 & 78.4 & 62.3 & 42.0 \\
         & $\text{AR}_{\text{FUDGE}\:\gamma=7}$       & 925  & 13  & 0.48 & 11.1 & 9.8  & 9.6  & 9.2  \\
         \cline{2-9}
         & MDLM                             & 596  & 271 & 0.45 & 85.9 & 73.7 & 61.1 & 44.0 \\
         & $\text{MDLM}_\text{(D-CFG: $\gamma=3$)}$      & 772  & 53  & 0.41 & 87.8 & 73.9 & 54.2 & 22.5 \\
         & $\text{MDLM}_\text{(D-CBG: $\gamma=3$)}$      & 436  & 21  & 0.37 & 50.5 & 48.6 & 46.1 & 44.7 \\
         \cline{2-9}
         & UDLM                            & 895  & \underline{345} & 0.47 & 89.4 & 88.0 & 58.1 & 37.8 \\
         & $\text{UDLM}_\text{(D-CFG: $\gamma=5$)}$      & 850  & 69  & 0.47 & 80.6 & 58.6 & 35.9 & 13.9 \\
         & $\text{UDLM}_\text{(D-CBG: $\gamma=10$)}$     & 896  & \textbf{ 374} & 0.47 & 90.1 & 77.8 & 58.6 & 37.7 \\
         \cline{2-9}
         & $\textbf{NSD}_{\tau=3.0}$          & 353  & 36  & \textbf{ 0.63} & \textbf{ 0.0} & \textbf{ 0.0}   & \textbf{ 0.0}   & \textbf{ 0.0}   \\
         & $\textbf{NSD}_{\tau=3.5}$          & 863  & 91  & \underline{0.62} & - & \textbf{ 0.0} & \textbf{ 0.0} & \textbf{ 0.0} \\
         & $\textbf{NSD}_{\tau=4.0}$          & 936  & 108 & 0.61 & - & -   & \textbf{ 0.0} & \textbf{ 0.0} \\
         & $\textbf{NSD}_{\tau=4.5}$          & 938  & 121 & 0.58 & - & -   & -   & \textbf{ 0.0} \\
         \cline{1-9}
\end{tabular}
}
\end{minipage}

\caption{Molecule generation constrained to strict synthetic accessibility thresholds.}
\label{fig:sa}
\end{table}

\subsection{Motion Planning for Autonomous Agents (Safety)}
\label{appendix:motion}

\begin{figure}[ht!]
\begin{minipage}{0.4\linewidth}
    \centering
    \includegraphics[width=0.85\columnwidth]{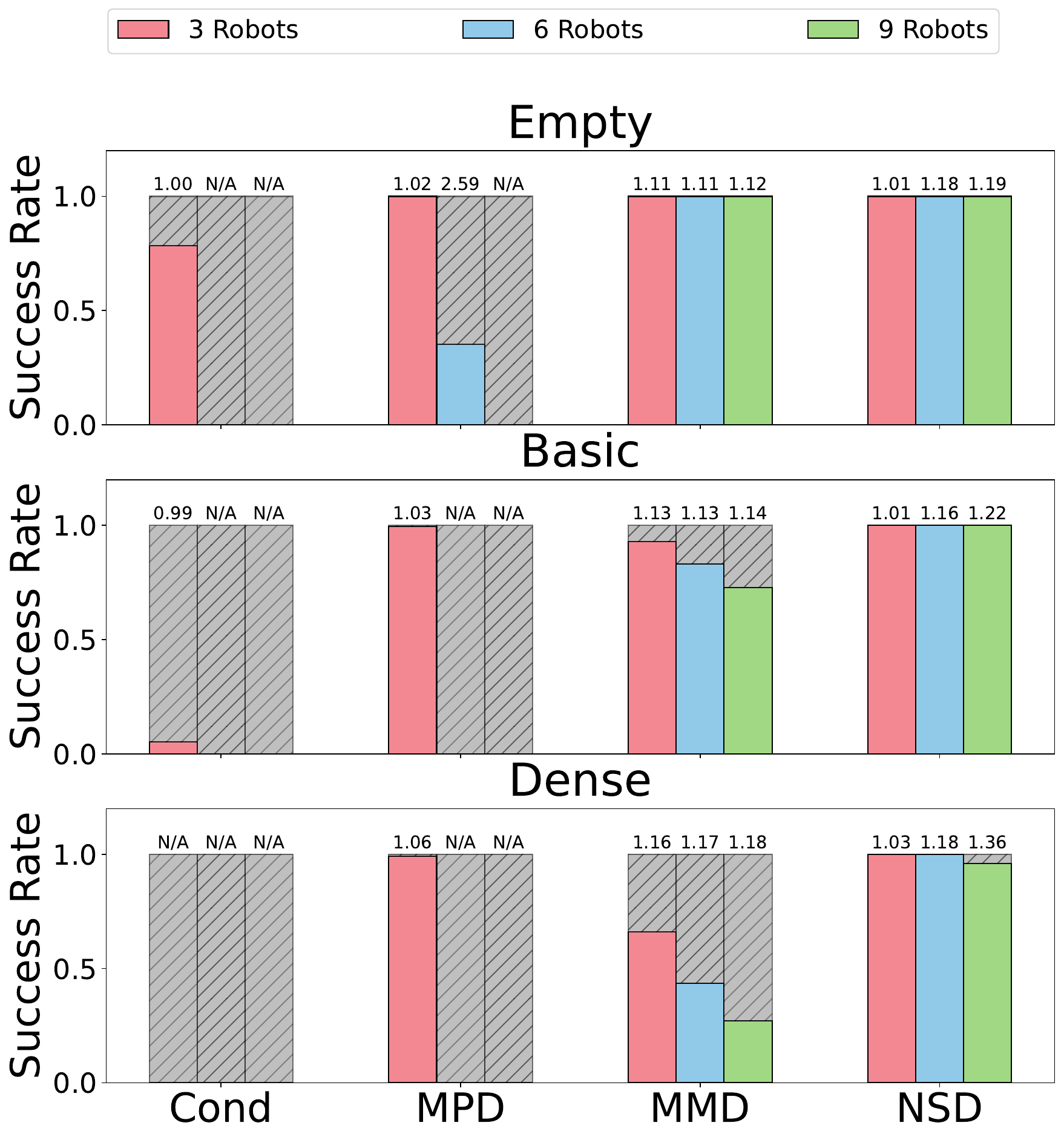}
    \label{fig:Results of Random Maps}
\end{minipage}
\hfill
\begin{minipage}{0.58\linewidth}
    \centering
    \subfigure[Basic Maps.]
    {
        \includegraphics[width=0.45\columnwidth]{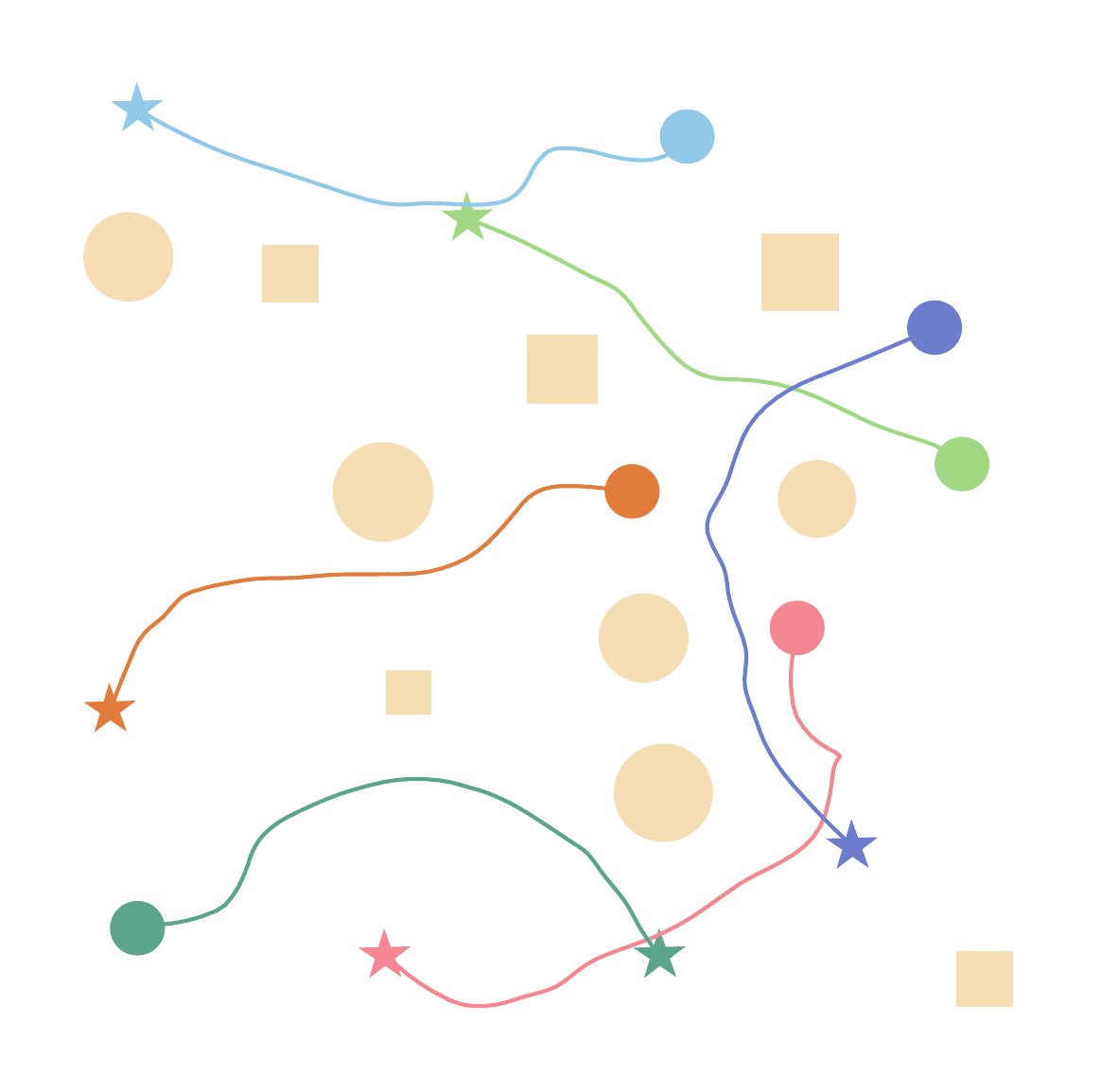}
        \label{fig:Trajectories for Basic Maps}
    }
    \subfigure[Dense Maps.]
    {
        \includegraphics[width=0.45\columnwidth]{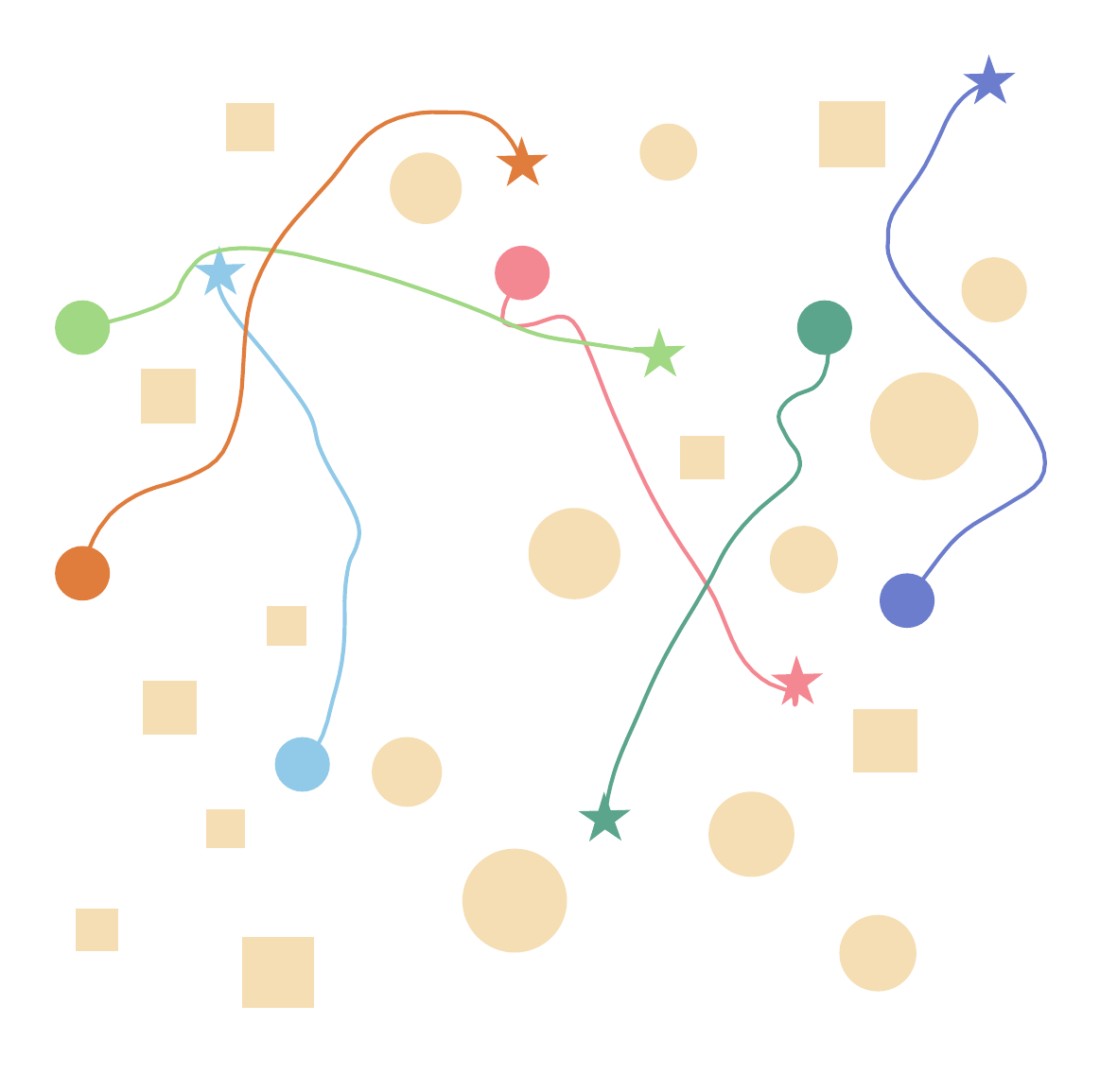}
        \label{fig:Trajectories for Dense Maps}
    }
    \label{fig:Trajectories of Random Maps}
\end{minipage}
\caption{Evaluation on random maps with three different numbers of robots. Gray bars represents the failure rate, and values on top of the bars indicate average path length per robot.}
\label{fig:random_maps}
\end{figure}

For this task, we begin by assigning start and goal states for a series of agents in each respective environment. For practical maps (Figure \ref{fig:motion_planning}), these positions fall in predefined zones that reflect real-world constraints, such as designated pickup and drop-off locations in a warehouse. For random maps (Figure \ref{fig:random_maps}), we assign these start and goal state randomly, constraining these to be collision-free locations, ensuring feasible solutions exist.
To further ensure this, we discretize the environments and apply multi-agent pathfinding algorithms to verify the existence of collision-free solutions. If no valid assignment can be found, we regenerate the configuration.

Our results are evaluated by success rate (bars included in the figures), the frequency at which state-of-the-art methods find feasible solutions, and path length (at the top of the bars), a metric for the optimality of the solutions. Experiments are conducted with three, six, and nine robots, generating ten test cases per configuration.
In addition to evaluation on practical map environments illustrated in Figure \ref{fig:motion_planning}, we provide additional evaluation on random maps in Figure \ref{fig:random_maps}. Again, we see that NSD dramatically outperforms the baselines in its ability to generate feasible motion trajectories. This is particularly exasperated as we scale the number of agents and obstacles. Of the baselines, only MMD is able to ever provide feasible solutions for nine agents, although we note that it has much more frequent constraint violations than NSD on non-empty maps.

\paragraph{Additional benchmarks.}
To supplement our evaluation, we compare to several domain specific approaches:
\begin{enumerate}[leftmargin=*, parsep=0pt, itemsep=0pt, topsep=0pt]
    \item \textbf{Conditional Diffusion Model (Cond)}: A matching diffusion model implementation to NCS, fine-tuned on benchmark trajectories to address autonomous motion planning problems \citep{nichol2021improved}.
    \item \textbf{Motion Planning Diffusion (MPD)}: The previous state-of-the-art for single-robot motion planning~\citep{carvalho2023motion}, this approach is extended to handle multi-agent settings for comparative analysis.
    \item \textbf{Multi-robot Motion Planning Diffusion (MMD)}: A recently proposed method that integrates diffusion models with classical search techniques, generating collision-constrained multi-agent path planning solutions~\citep{shaoul2024multi}.
\end{enumerate}

\paragraph{Symbolic test.}
We model two types of collision constraint tests. The first, \(\phi_1\) corresponds to collisions between agents, whereas \(\phi_2\) captures collisions between agents and obstacles in the map. 
We can express the collision-avoidance constraints as follows. First, for inter-agent separation, we require that for every pair of distinct agents \(i\) and \(i'\) and at every time step \(j\), their positions are separated by at least a minimum distance \(d_{\min}\) (which is defined as the sum of their radii):

\[
\phi_1(\bm{x}) \defeq \quad \forall i, i' \ (i \neq i'),\; \forall j \quad \|\bm{p}_i^j - \bm{p}_{i'}^j\|_2 \ge d_{\min}.
\]
Second, to ensure agents do not collide with obstacles, we require that for each agent \(i\) at each time step \(j\) and for every obstacle \(k\) with radius \(r_k\), the agent’s position is at least \(\bm{r}_k\) away from the obstacle’s center \(\bm{o}_k\):
\[
\phi_2(\bm{x}) \defeq \quad \forall i,\; \forall j,\; \forall k \quad \|\bm{p}_i^j - \bm{o}_k\|_2 \ge \bm{r}_k.
\]
Here, \(i\) and \(i'\) index the agents, \(j\) denotes the time steps, \(\bm{p}_i^j\) is the position of agent \(i\) at time \(j\), and \(\bm{o}_k\) is the position of obstacle \(k\). As these constraints can be expressed in closed-form, we model \(\Delta\phi\) directly from this, employing the augmented Lagrangian method to solve this projection (due to the highly non-convex nature of these constraints).

\subsection{Morphometric Property Specification (Data Scarcity and Domain Generalization)}
\label{appendix:microstructure}

As mentioned in the main text, our dataset is generated by subsampling a single 3,000\(\times\)3,000 pixel image into patches of size 64\(\times\)64. We then upscale these patches to 256\(\times\)256 images to increase the resolution for our generation task. The data is obtained from \cite{chun2020deep}, and we reiterate it contains only a small range of porosity values fall within the desired porosity ranges (e.g., at the porosity level P(\%) 50, only 7\% of the training data falls within a generous five percent error margin to either side), contributing to the challenging nature of this setting.

In the analysis of both natural and synthetic materials, heuristic metrics are commonly used to quantify microstructure features such as crystal shapes and void distributions. These measures provide qualitative insights into the fidelity of the synthetic samples relative to the training data. Here, we present the distributions of three microstructure descriptors, following the approach of \citeauthor{chun2020deep}

The results demonstrate that the explicit constraint enforcement in NSD yields microstructures that more faithfully replicate the ground truth. In contrast, the conditional model tends to produce certain features at frequencies that do not align with the training distribution. By integrating porosity and related constraints during the sampling process, NSD is able to generate a set of microstructures that is both more representative and accurate.

\begin{figure}[ht!]
    \centering
    \includegraphics[width=0.99\linewidth]{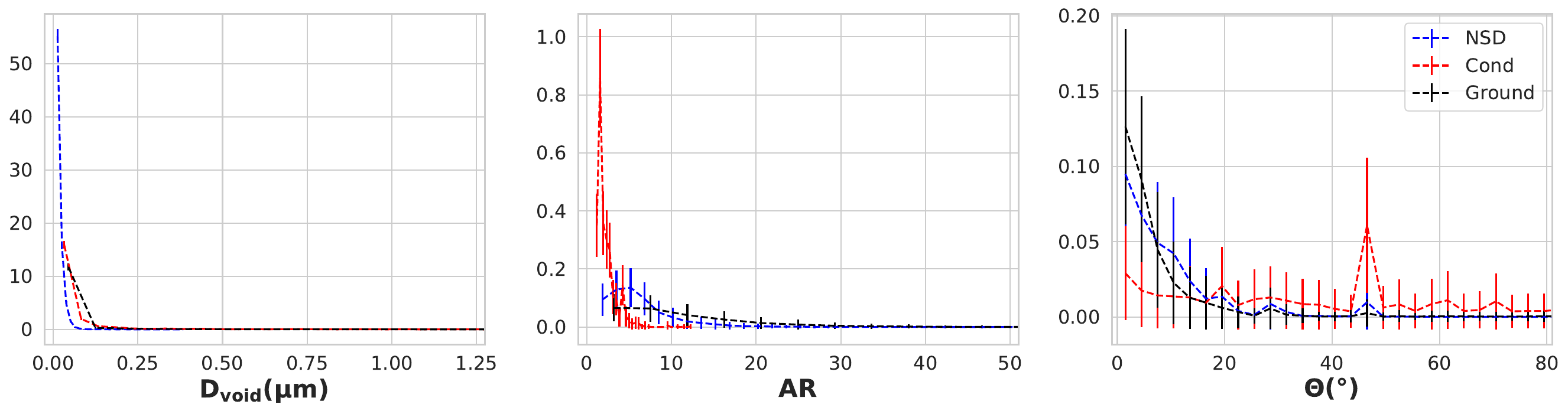}
    \caption{Morphometric parameter distributions comparing ground truth microstructures with those generated by the \textsl{\bcol{NSD}} and \textsl{Cond} models, evaluated using heuristic analysis.}
    \label{fig:morph_heuristic}
\end{figure}

\paragraph{Additional benchmarks.}
To supplement our evaluation, we compare to several domain specific approaches:
\begin{enumerate}[leftmargin=*, parsep=0pt, itemsep=0pt, topsep=0pt]
    \item \textbf{Conditional Diffusion Model (Cond)}: A conditional diffusion model implementation modeled from \cite{chun2020deep}.
    \item \textbf{Post-Processing ($\text{Post}^+$)}: A matching implementation to our diffusion model for NSD, with the projection steps omitted from the sampling process, except after the final step.
    \item \textbf{Conditional + Post-Processing ($\text{Cond}^+$}): The Cond model, but with the addition of a post-processing projection after the final step.
\end{enumerate}

\paragraph{Symbolic test.}
We define a test function \(\phi\) that measures the porosity of an image:
\[
\phi(\bm{x}) \defeq \bigl( \;\sum_{i=1}^n \sum_{j=1}^m  \bigl(\bm{x}^{i,j} < 0\bigr) \;\bigr) = K,
\]
where \(\bm{x}^{i,j} \in [-1,1]\) is the pixel value at row \(i\) and column \(j\). Our desired constraint is that the porosity of the generated image \(\bm{x}\) must equal a target value \(K\).
In our framework, this condition is used as a test that triggers the projection: if \(\phi(\bm{x}) \neq 1\), a projection operator is applied to minimally adjust \(\bm{x}\) so that the constraint is satisfied. 
This can be constructed using a top-k algorithm to return,
\begin{subequations}
\begin{align*}
    \mathcal{P}_\mathbf{C}(\bm{x}) = \arg\min_{\bm{y}^{i,j}} \sum_{i,j} \|\bm{y}^{i,j} - \bm{x^{i,j}} \| \\
    \text{s.t. } \quad \bm{y}^{i,j} \in [-1, 1], \quad \sum^n_{i=1} \sum^m_{j=1} \left( \bm{y^{i,j}} < 0\right) = K
\end{align*}
\end{subequations}
where $K$ is the number of pixels that should be ``porous''.
Because this program is convex, it serves as a certificate that the generated images comply with the prescribed porosity constraint.

\subsection{Physics-informed Motion (Data Scarcity and Domain Generalization)}
\label{appendix:phys}
\begin{figure}[ht]
  \centering
  \begin{minipage}{0.59\textwidth}
    \centering
    \setlength{\tabcolsep}{1pt}
    \ra{0.25}
    \vspace{20pt}
    \begin{tabular}{c cccc c cccc}
      \toprule
      \multirow{2}{*}{t} & \multicolumn{4}{c}{\footnotesize{ Earth (in distribution)}} & ~ & \multicolumn{4}{c}{\footnotesize{ Moon (out of distribution)}}\\
      \cline{2-5} \cline{7-10}
      & \footnotesize{\sl Ground} & \footnotesize{\sl \bcol{NSD}} & \footnotesize{\sl Post$^+$} & \footnotesize{\sl \rcol{Cond$^+$}} &
      & \footnotesize{\sl Ground} & \footnotesize{\sl \bcol{NSD}} & \footnotesize{\sl Post$^+$} & \footnotesize{\sl \rcol{Cond$^+$}} \\
      \midrule
      {1} & \includegraphics[width=0.11\textwidth]{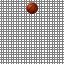} 
          & \includegraphics[width=0.11\textwidth]{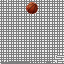} 
          & \includegraphics[width=0.11\textwidth]{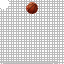} 
          & \includegraphics[width=0.11\textwidth]{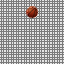} & 
          & \includegraphics[width=0.11\textwidth]{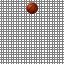} 
          & \includegraphics[width=0.11\textwidth]{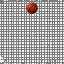} 
          & \includegraphics[width=0.11\textwidth]{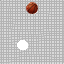} 
          & \includegraphics[width=0.11\textwidth]{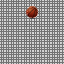} \\
      {3} & \includegraphics[width=0.11\textwidth]{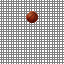} 
          & \includegraphics[width=0.11\textwidth]{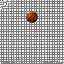} 
          & \includegraphics[width=0.11\textwidth]{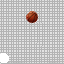} 
          & \includegraphics[width=0.11\textwidth]{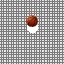} &
          & \includegraphics[width=0.11\textwidth]{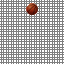} 
          & \includegraphics[width=0.11\textwidth]{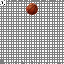} 
          & \includegraphics[width=0.11\textwidth]{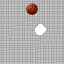} 
          & \includegraphics[width=0.11\textwidth]{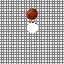} \\
      {5} & \includegraphics[width=0.11\textwidth]{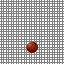} 
          & \includegraphics[width=0.11\textwidth]{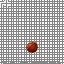} 
          & \includegraphics[width=0.11\textwidth]{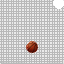} 
          & \includegraphics[width=0.11\textwidth]{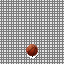} &
          & \includegraphics[width=0.11\textwidth]{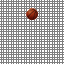} 
          & \includegraphics[width=0.11\textwidth]{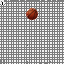} 
          & \includegraphics[width=0.11\textwidth]{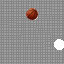} 
          & \includegraphics[width=0.11\textwidth]{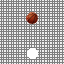} \\
      \bottomrule
    \end{tabular}
    \vspace{20pt}
    \begingroup
    \renewcommand{\arraystretch}{0.8}
      \setlength{\tabcolsep}{6.2pt}
      \hspace{-8pt}
      \begin{tabular}{rccccc}
        & \footnotesize{\sl \bcol{NSD}} & \footnotesize{\sl Post$^+$} 
        & \footnotesize{\sl \rcol{Cond}} & \footnotesize{\sl \rcol{Cond$^+$}} \\
        \midrule
        \footnotesize{\textbf{FID:}} & \footnotesize{26.5 $\pm$ 1.7} & \footnotesize{52.5 $\pm$ 1.0} 
        & \footnotesize{\textbf{22.5 $\pm$ 0.1}} & \footnotesize{53.0 $\pm$ 0.3} \\
        \footnotesize{\textbf{Viol. (\%):}} & \footnotesize{\textbf{0.0}} & \footnotesize{\textbf{0.0}} 
        & \footnotesize{100.0} & \footnotesize{\textbf{0.0}} \\
        \bottomrule
      \end{tabular}
    \endgroup
  \end{minipage}%
  \hfill
  \begin{minipage}{0.38\textwidth}
    \centering
    \includegraphics[width=\textwidth]{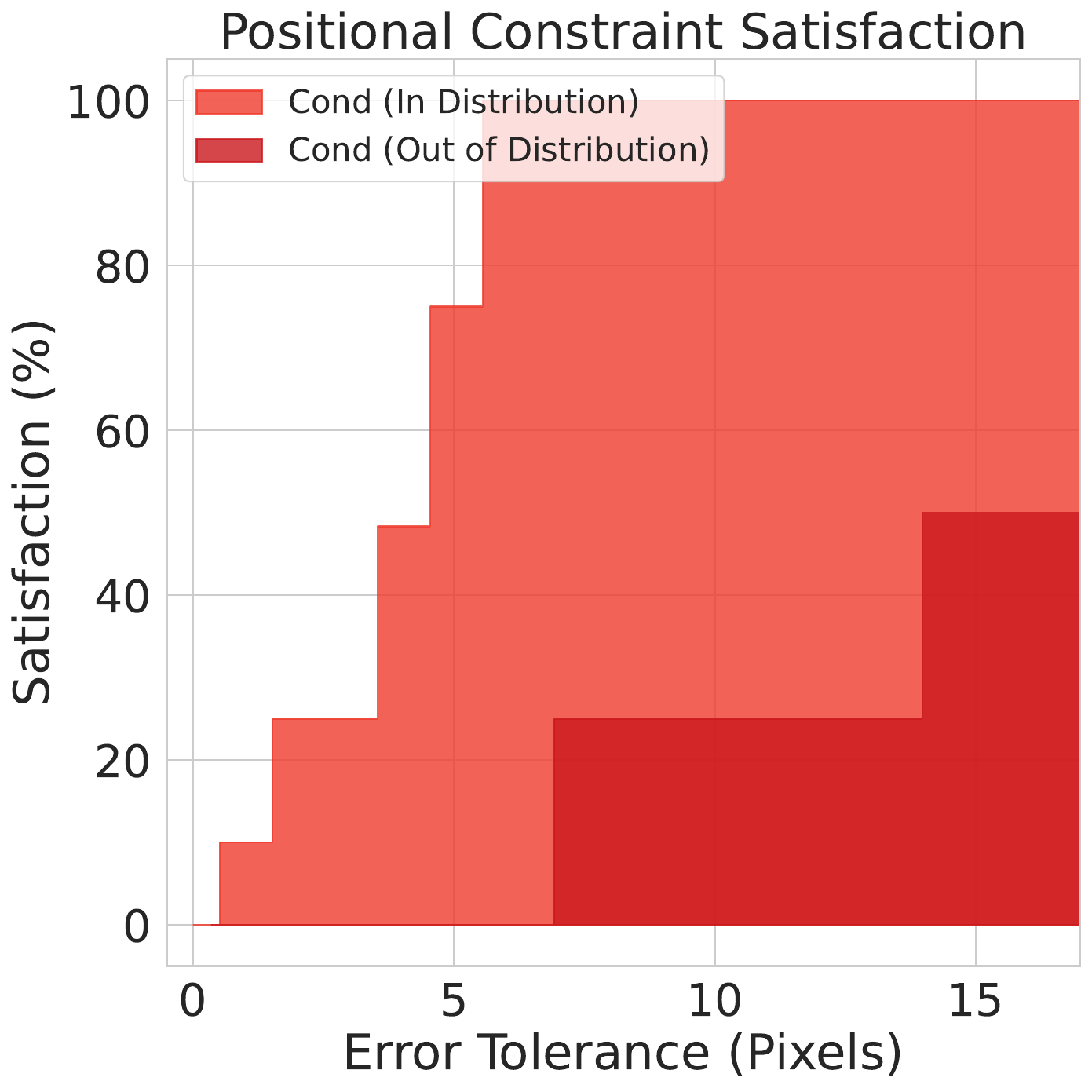}
  \end{minipage}
  \caption{Physics-informed motion experimental results.}
  \label{fig:morph_phys}
\end{figure}

For this setting, we generate a series of video frames depicting an object accelerating due to gravity.
Here, the object’s motion is governed by a system of ordinary differential equations (Eq.~\eqref{eq:ball-position}), which our method directly integrates into the constraint set (\(\phi\)). 
In addition to the complexity of the constraints, two key challenges are posed: \textbf{(1) Data Scarcity:} Our training data is based solely on Earth's gravity, yet our model is tested on gravitational forces from the Moon and other planets, where \emph{there are no feasible training samples provided}, and, consequentially, \textbf{(2) Out-of-Distribution Constraints:} the imposed constraints are not represented in the training.

Figure~\ref{fig:morph_phys} (left) highlights the results of our experiments;
standard conditional diffusion models often produce objects that are misplaced within the frame, as evidenced by white object outlines in the generated samples and the reported constraint violations on the right side of the figure. 
Post-processing approaches correct positioning at the cost of significant image degradation.
In contrast, our method \textbf{guarantees} \emph{satisfaction of physical constraints while maintaining high visual fidelity}, producing samples that fully satisfy positional constraints. These results demonstrate that our approach generalizes to out-of-distribution physical conditions while ensuring strict compliance with governing physical laws.

For training, we begin by generating a dataset uniformly sampling various object starting points within the frame size [0,63]. For each data point, six frames are produced, depicting the objects movement as governed by the ODE in Equation \eqref{eq:ball-position}. The velocity is initialized to \(\mathbf{v}_0=0\). The diffusion models are trained on 1000 examples from this dataset, using a 90/10 training and testing split. 
The conditional model is implemented following \cite{voleti2022mcvd}, where we provide two frames illustrating the motion as conditioning. The model then infers future frames from these to produce the final videos.

\paragraph{Additional benchmarks.}
To supplement our evaluation, we compare to several domain specific approaches:
\begin{enumerate}[leftmargin=*, parsep=0pt, itemsep=0pt, topsep=0pt]
    \item \textbf{Conditional Diffusion Model (Cond)}: A conditional diffusion model implementation as outlined by \cite{voleti2022mcvd}.
    \item \textbf{Post-Processing ($\text{Post}^+$)}: A matching implementation to our diffusion model for NSD, with the projection steps omitted from the sampling process, except after the final step.
    \item \textbf{Conditional + Post-Processing ($\text{Cond}^+$}): The Cond model, but with the addition of a post-processing projection after the final step.
\end{enumerate}

\paragraph{Symbolic test.}
We define a test function \(\phi\) that checks whether the object's position in a frame meets the prescribed positional constraints given by Equations:

\begin{subequations}
\label{eq:ball-position}
\begin{minipage}[b]{0.50\linewidth}
    \begin{align}
        \mathbf{p}_{t} &= \mathbf{p}_{t-1} + \left(\mathbf{v}_t   + \left(0.5 \times \frac{\partial \bm{v}_{t}}{\partial t}\right)\right)
        \label{eq:pt}
    \end{align}
\end{minipage}
\begin{minipage}[b]{0.40\linewidth}
    \begin{align}
    \mathbf{v}_{t+1} &= \frac{\partial \bm{p}_{t}}{\partial t} + \frac{\partial \bm{v}_{t}}{\partial t},
    \label{eq:vt}
    \end{align}
\end{minipage}
\vspace{4pt}
\end{subequations}\\
Hence, we define our test function:
\[
\phi(\bm{x}) \defeq \bigl(\text{object position in } \bm{x} \text{ equals } \mathbf{p}_t)\bigr.
\]
In other words, the generated frame \(\bm{x}\) is considered feasible if the object's position exactly matches the target position \(\mathbf{p}_t\) computed by the dynamics model.

When \(\phi(\bm{x}) \neq 1\), a projection operator is triggered to enforce the positional constraint. This projection proceeds in two steps. First, the object's current location is detected and its pixels are set to the maximum intensity (i.e., white), effectively removing the object from its original position while storing the indices of the object’s structure. Second, the object is repositioned by mapping its stored pixel indices onto the center point corresponding to \(\mathbf{p}_t\). If the frame already satisfies the positional constraint (i.e., \(\phi(\bm{x}) = 1\)), the projection leaves the image unchanged. Since this projection process is well-defined and convex, it provides a certificate that the generated frames comply with the prescribed positional constraints.

\subsection{Safe Text Generation (Safety)}
\label{appendix:safe_text}

As language models become widely adopted for commercial and scientific applications, it is necessary that generated text adheres to safety constraints, preventing 
the production of harmful or toxic content. 
Hence, it is essential to provide \textbf{(1) Safety-Critical Outputs:} the adoption of constraint-aware methods is essential for these applications, especially considering recent examples of \emph{toxic outputs encouraging self-harm or providing information which could be used to harm others} \citep{perez2022red}.

Table~\ref{tab:sent_toxic} highlights the results of our experiments, which evaluate language models on toxicity mitigation using prompts from the RealToxicityPrompts dataset \citep{gehman2020realtoxicityprompts}. Our method significantly improves control over generated content by enforcing strict toxicity constraints during inference. Compared to baseline models, 
such as GPT-2 and Llama 3.2, 
which exhibit high violation rates, \emph{NSD achieves perfect constraint satisfaction across all toxicity thresholds}.
Furthermore, our approach scales effectively, ensuring robust toxicity mitigation even at increasingly strict thresholds (denoted as \(\tau\)).

\begin{table}[ht!]
  \centering
  \begin{tabular}{|l|lr|rr|rrr|}
  \hline
  \multirow{9}{*}{\rotatebox{90}{\footnotesize \textbf{Sentence Toxicity}}}%
    & \multirow{2}{*}{\textbf{Model}} 
    & \multirow{2}{*}{\textbf{Size}} 
    & \multicolumn{2}{c|}{\textbf{PPL}} 
    & \multicolumn{3}{c|}{\textbf{Viol (\%)}} \\
    &  &  & \textbf{Mean} & \textbf{Median}
       & \(\tau=0.25\) & \(\tau=0.50\) & \(\tau=0.75\) \\
  \cline{2-8}
    & GPT2 & 124M 
       & 19.1 & 17.6
       & 36.3 & 23.8 & 16.2 \\
    & $\text{GPT2}_\textit{PPLM}$ & 345M 
       & 47.6 & 37.2
       & 15.2 & 8.1  & 4.3  \\
    & $\text{GPT2}_{\text{FUDGE}_{\lambda=2}}$ & 124M
   & 26.46 & 18.79  & 31.5 & 19.7 & 12.7 \\
   & $\text{GPT2}_{\text{FUDGE}_{\lambda=9}}$ & 124M
   & 81.84 & 19.22 & 30.6 & 19.6 & 11.7 \\
    & $\text{Llama 3.2}$ & 1B 
       & \textbf{15.7} & \textbf{14.6}
       & 34.9 & 27.8 & 23.1 \\
    & $\text{MDLM}$  & 110M 
       & 46.7 & 39.8
       & 32.1 & 23.2 & 17.2 \\
    & $\textbf{NSD}_{\tau=0.25}$ (Ours) & 110M 
       & 61.6 & 45.4
       & \textbf{0.0} & \textbf{0.0} & \textbf{0.0} \\
    & $\textbf{NSD}_{\tau=0.50}$ (Ours) & 110M 
       & 59.4 & 44.2
       & -- & \textbf{0.0} & \textbf{0.0} \\
    & $\textbf{NSD}_{\tau=0.75}$ (Ours) & 110M 
       & 54.9 & 43.2
       & -- & -- & \textbf{0.0} \\
  \hline
  \end{tabular}
  \caption{Results for safe text generation at various toxicity levels \(\tau\).}
  \label{tab:sent_toxic}
\end{table}

Among the baselines, GPT-2 (124M) and LLaMA (1B) achieve the lowest perplexity scores; however, they frequently generate toxic content, leading to high violation rates. While GPT-2 + PPLM (345M) \cite{dathathri2019plug} reduces violations across all toxicity thresholds, it fails to consistently prevent toxic generations and suffers from increased perplexity. MDLM (110M) exhibits higher perplexity than GPT-2, with a median PPL of 39.8, and although it moderately reduces toxicity violations compared to GPT-2, the rates remain significant. In contrast, NSD achieves \emph{perfect constraint satisfaction} across all toxicity thresholds while maintaining sentence fluency.

\paragraph{Additional benchmarks.}
To supplement our evaluation, we compare to several domain specific approaches:
\begin{enumerate}[leftmargin=*, parsep=0pt, itemsep=0pt, topsep=0pt]
    \item \textbf{GPT2}: Our model uses a GPT2 tokenizer and is roughly the same size as GPT2, so we add this as a point of comparison.
    \item \textbf{Llama 3.2}: For comparison to state-of-the-art autoregressive models, we employ Llama 3.2, noting that this model is an order of magnitude larger than our diffusion model.
    \item \textbf{Plug and Play Language Model (\(\text{GPT2}_\textit{PPLM}\))}: We utilize gradient-based guidance as proposed by \cite{dathathri2019plug} to condition autoregressive generation against producing toxic outputs.
    \item \textbf{Masked Diffusion Model ($\text{MDLM}$)}: A masked discrete language diffusion model implementation from \cite{schiff2024simple}.
\end{enumerate}

\paragraph{Symbolic test.}
As \(\phi\) cannot be explicitly modeled for general text toxicity quantification, we train a surrogate model to provide a differentiable scoring metric \(\Delta\phi\).
Hence, the constraint is assessed with respect to this learned metric, such that:
\(
 \Delta\phi(\bm{x}^\star) \leq \tau,
\)
where \(\tau\) is a tunable threshold that controls the degree of toxicity that's permissible (lower values resulting in less toxic output sequences).
As the surrogate model for toxicity task, we use a GPT-Neo (1.3B) model, adapted for binary classification. We finetune this model on the Jigsaw toxicity dataset which includes multiple toxicity-related labels such as toxic, severe toxic, insult, etc. We consolidates these columns into a single binary target (toxic vs. non-toxic).

\section{Missing Proofs}
\label{appendix:theory}

\begin{proof}\textbf{(Theorem~\ref{thrm:conv})}
By optimization theory of convergence in a convex setting, provided an arbitrary number of update steps $t$, $\bm{x}_{t}$ will reach the global minimum. Hence, this justifies the existence of $\bar{t}$ as at some iteration as $T \xrightarrow{} \infty$, 
\[
\|\mathcal{U}_\theta(\bm{x}_t) - \Phi\|_2 \leq \|\rho - \Phi\|_2 
\]
which will hold for every iteration thereafter.

\end{proof}

\begin{proof}\textbf{(Theorem~\ref{eq:theorem-main})}
For any update taken after convergence, consider a gradient update without the stochastic noise. There are two cases:

\paragraph{Case 1:} Suppose that
\(
\mathcal{U}_\theta(\bm{x}_t)
\)
is closer to the optimum than \(\rho\). By the definition of \(\rho\), this implies that \(\bm{x}_t\) is infeasible. Moreover, a gradient step taken from an infeasible point will yield an update that is closer to the optimum than any point achievable from the feasible set. Hence, we obtain:
\begin{equation}
\label{eq:error-ineq}
\textit{Error}\Bigl(\mathcal{U}_\theta(\bm{x}_t)\Bigr) >  
\textit{Error}\Bigl(\mathcal{P}_{\mathbf{C}}\bigl(\mathcal{U}_\theta(\bm{x}_t)\bigr)\Bigr).
\end{equation}

\paragraph{Case 2:} Suppose instead that 
\(
\mathcal{U}_\theta(\bm{x}_t)
\)
is equally close to the optimum as \(\rho\). In this situation, either (1) \(\bm{x}_t\) is already the closest feasible point to the optimum (i.e., \(\bm{x}_t = \mathcal{P}_{\mathbf{C}}(\bm{x}_t)\)), so that the error terms are equal, or (2) \(\bm{x}_t\) is infeasible. In the latter case, the gradient step from \(\bm{x}_t\) is equivalent in magnitude to that from the nearest feasible point, but, by convexity, the triangle inequality ensures that the error from starting at an infeasible point exceeds that from starting at the feasible projection. Thus, Equation~\eqref{eq:error-ineq} holds in all cases.
Finally, when the stochastic noise (sampled from a zero-mean Gaussian) is incorporated, taking the expectation over the update yields:
\begin{align*}
\label{eq:error-final}
\mathbb{E} \Bigl[ \textit{Error}\Bigl(\mathcal{U}_\theta(\bm{x}_t)\Bigr) \Bigr] \geq 
\mathbb{E} \Bigl[ \textit{Error}\Bigl(\mathcal{P}_{\mathbf{C}}\bigl(\mathcal{U}_\theta(\bm{x}_t)\bigr)\Bigr) \Bigr].
\end{align*}

\end{proof}

\end{document}